\documentclass{article}

\usepackage{arxiv}

\usepackage[utf8]{inputenc} 
\usepackage[T1]{fontenc}    
\usepackage{hyperref}       
\usepackage{url}            
\usepackage{booktabs}       
\usepackage{amsfonts}       
\usepackage{nicefrac}       
\usepackage{microtype}      
\usepackage{lipsum}		    
\usepackage{graphicx}
\usepackage{authblk}
\usepackage{physics}
\usepackage{multirow}
\usepackage{amsmath}
\usepackage{pgfplots}
\usepgfplotslibrary{statistics}
\usepackage{pgfplotstable}
\pgfplotsset{compat=1.18}
\usepackage{pgfplots}
\usepackage{filecontents}
\usepackage{svg}
\usepackage{booktabs}
\usepackage{subcaption}
\usepackage{comment}
\usepackage{caption}
\usepackage{cite}
\usepackage{xcolor}
\usepackage{cancel}
\usepackage{fancyhdr}
\definecolor{darkblue}{RGB}{0, 51, 102}     
\definecolor{mediumblue}{RGB}{51, 102, 153} 
\definecolor{lightblue}{RGB}{153, 204, 255} 

\definecolor{darkgreen}{RGB}{0, 102, 51}    
\definecolor{mediumgreen}{RGB}{51, 153, 102} 
\definecolor{lightgreen}{RGB}{153, 200, 160} 

\definecolor{darkred}{RGB}{102, 0, 0}       
\definecolor{mediumred}{RGB}{153, 51, 51}   
\definecolor{lightred}{RGB}{255, 153, 153}  

\definecolor{darkpurple}{RGB}{75, 0, 130}     
\definecolor{mediumpurple}{RGB}{153, 50, 204} 
\definecolor{lightpurple}{RGB}{170, 110, 170} 

\newcommand{\bs}[1]{\boldsymbol{#1}}

\svgsetup{inkscapelatex=false}

\title{Thermodynamics-Informed Graph Neural Networks for Real-Time Simulation of Digital Human Twins}

\author[1]{Lucas Tesán}
\author[1]{David González}
\author[1,2]{Pedro Martins}
\author[1]{El\'ias Cueto}
\affil[1]{{\small ESI Group-UZ Chair of the National Strategy on Artificial Intelligence. \protect\\ Aragon Institute of Engineering Research (I3A). Universidad de Zaragoza. Zaragoza, Spain.}}
\affil[2]{{\small Aragonese Foundation for Research and Development (ARAID), Aragón, Spain. }}

\begin{document}
\maketitle

\begin{abstract}

	The growing importance of real-time simulation in the medical field has exposed the limitations and bottlenecks inherent in the digital representation of complex biological systems. This paper presents a novel methodology aimed at advancing current lines of research in soft tissue simulation. The proposed approach introduces a hybrid model that integrates the geometric bias of graph neural networks with the physical bias derived from the imposition of a metriplectic structure as soft and hard constrains in the architecture, being able to simulate hepatic tissue with dissipative properties. This approach provides an efficient solution capable of generating predictions at high feedback rate while maintaining a remarkable generalization ability for previously unseen anatomies.  This makes these features particularly relevant in the context of precision medicine and haptic rendering.
 
    Furthermore, this work synthesizes two prominent concepts in recent years: the role of message passing as a geometric mechanism fundamental to graph neural networks, and the potential of thermodynamics-informed networks to enhance extrapolation capabilities beyond training scenarios. We develop a multi-graph interaction between the computational model of the liver and a surgical tool. A displacement imposed at the contact region initiates a controlled flow of information that propagates throughout the graph model, aiming to achieve a steady and more efficient exchange of information across the entire network. The physics bias is obtained by imposing a metriplectic structure, enforced via strong and soft constraints. This ensures that the network satisfies thermodynamic principles during inference, even for a previously unseen system.
    
    Based on the adopted methodologies, we propose a model that predicts human liver responses to traction and compression loads in as little as 7.3 milliseconds for optimized configurations and as fast as 1.65 milliseconds in the most efficient cases, all in the forward pass. The model achieves relative position errors below 0.15\%, with stress tensor and velocity estimations maintaining relative errors under 7\%. This demonstrates the robustness of the approach developed, which is capable of handling diverse load states and anatomies effectively. This work highlights the feasibility of integrating real-time simulation with patient-specific geometries through deep learning, paving the way for more robust digital human twins in medical applications. 

\end{abstract}


\newpage

\section{Introduction}
\label{sec:intro}

In recent decades, real-time simulation has emerged as a fundamental tool for modeling complex systems in diverse fields such as engineering, biology, physics, and social sciences. These systems, which range from structural mechanics \cite{ritto2021} to climate models \cite{lam2022graphcast} and even biological systems themselves \cite{bruynseels2018,coorey2022health,BlancaDTw}, require significant computational power to ensure accuracy and efficiency in real-time decision-making.

In the context of this work, the primary focus is on the computational mechanics involved in modeling the response of a soft tissue, whose viscous-hyperelastic characteristics have long posed a classical challenge for established numerical methods \cite{marchesseau2017, martinez2014, TorneroCostaMartinezSanchis}. The issue lies in the mesh dependency of these methodologies \cite{zhou2004dynamic, ambati2011fem}, where both error and computational time are nonlinearly dependent on the element size, order, and shape factor, significantly constraining the model’s quality in environments that require high feedback rates.

Thus, the numerical methodology is based on iterative convergence towards the real solution, defined by the physical laws governing the model, and it is only limited by the spatio-temporal discretizations.  In contrast, data-driven approaches, usual in general machine learning, achieve convergence towards ground truth values largely based on the amount and diversity of data used for training. Under these conditions, inference is heavily influenced by the hyperspace of conditions defined in the training set, with hyperparameter tuning of the network playing a secondary, but relevant, role in determining the model performance.

As a result of these problems, unpredictability of so called \emph{black-box} dense models, new lines of research have emerged that propose alternatives to mitigate this pronounced bias, constraining the variability of possible solutions to those that best represent the physical reality defined by the underlying problem: geometric bias, widely popularized through the use of convolutional neural networks (CNNs) \cite{li2021survey}, defines and controls the mathematical structure of the optimized solution by leveraging the geometric properties of the data \cite{bronstein2017geometric}. In the case of CNNs, this translates into the localization of the pixels that make up an image and their positional relationships. In recurrent networks, the geometric bias manifests in the temporal dependency between the data \cite{medsker2001recurrent}, while in graph networks, a prior relationship is established before training. This prior describes the direction and extent of the information flow, although the exact effect of this connectivity is unknown \cite{pfaff2021learningmeshbasedsimulationgraph, wu2020comprehensive}.

On the other hand, physical biases assert that the data follow a differential structure governed by the physical laws that regulate the modeled process. These biases allow us  to establish weak constraints during loss optimization so that the residual of a differential equation or a degeneration term is satisfied, consequently aligning the model with the underlying physical principles and defining the so-called physics-informed neural networks (PINNs)\cite{raissi2019physics}.

\begin{figure}[h!]
    \centering
    \includegraphics[scale=0.6]{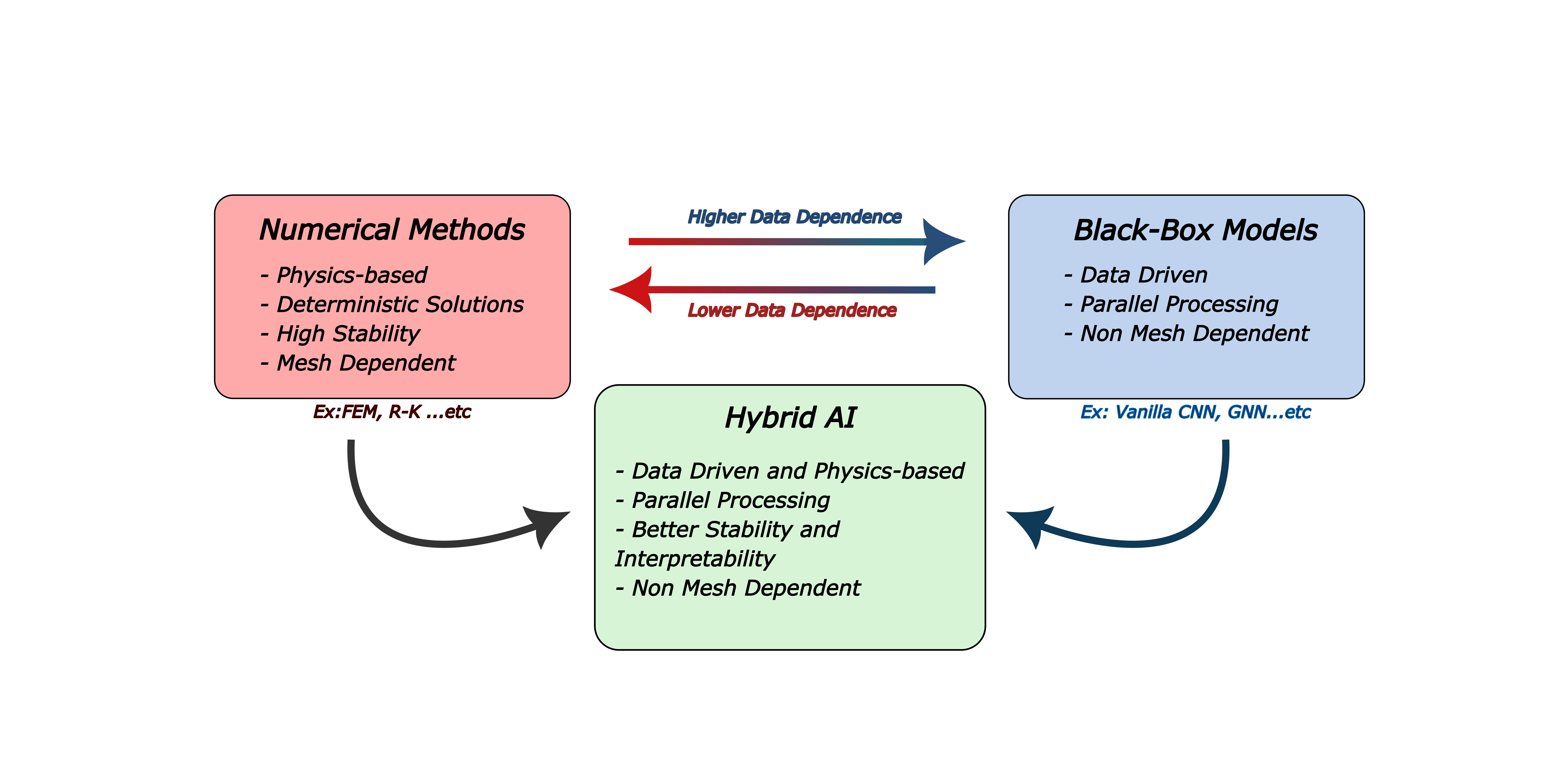}
    \caption{Comparison between the main characteristics of numerical methods, hybrid AI, and black box models.}
    \label{fig:HybridAI}
\end{figure}

In this way, when all this prior human knowledge translated into learning biases strengthens the predictive capacity of the network, a hybrid artificial intelligence is generated \cite{kamar2016directions}, forming the conceptual backbone of the current work, as shown in Figure \ref{fig:HybridAI}. 

\subsection{Objective}

The main objective of this work is to develop a robust hybrid neural simulator for real-time inference of the response of a patient-specific hepatic twin. As additional goals we also consider the validation of physical inductive biases through comparative analysis with data-only models. Additionally, it seeks to optimize convergence in message-passing processes to enhance response speed, ensuring a comprehensive flow of information across all nodes within the network's latent space. Finally, this study will analyze the model sensitivity in scenarios characterized by reduced training datasets and scarce data.

\subsection{Previous work} 

As we have pointed out, PINNs are a robust methodology for tackling systems whose governing partial differential equations are known. However, in many cases, this information is not available, or at least not completely, especially considering the natural complexity of biological systems. In the absence of knowing a system's governing equations, the establishment of a parallel between the learning of the physical phenomenon and that of a dynamical system is of great help in formulating the problem as a non-linear regression, that will provide an expression for the time evolution of the system under study \cite{weinan2017proposal}.

This approach enables the incorporation of specific physical properties into the learning process, ensuring compliance with natural first principles. For example, it allows learned simulators to uphold energy conservation within their predictions, an ability that has sparked considerable interest and was thoroughly investigated. Many studies, particularly on Hamiltonian \cite{Greydanus2019, Galimberti2023} and Lagrangian neural networks \cite{Bhattoo2023, Cranmer2020}, leverage these frameworks to ensure energy conservation during inference. However, a primary limitation of this method is that most physical processes are dissipative and irreversible at larger scales, requiring further conditions to satisfy both the first and second laws of thermodynamics.

Our work advances this field by introducing a novel approach that combines deep learning, mesh-based simulations, and physics-informed inductive biases, exploiting recent advancements  in these fields \cite{TIGNNs, hernandez2020structure}. Drawing on principles from geometric deep learning, we develop a method that models a system's physical interactions with external loads within a multi-graph framework \cite{pfaff2021learningmeshbasedsimulationgraph}, enforcing thermodynamic consistency through specialized inductive biases. This framework facilitates the hybrid learning of the dynamics from diverse physical systems, while ensuring spatial equivariance and scalability across varying resolutions \cite{HERNANDEZ2023115912}. While previous applications of graph neural networks to the neural simulation of biological systems exist, to the best of our knowledge, none of them employs physics-based inductive biases \cite{PEGOLOTTI2024107676}.

Furthermore, recent works that consider the region represented by each node as an open system are taken into account \cite{Tierz2024b}, employing a modification of the metriplectic formalism based on the port-Hamiltonian formalism \cite{Hern_ndez_2023Port}. This approach facilitates the development of a robust framework for the real-time simulation of soft tissue dynamics, establishing an innovative pathway for computing the mechanical response of the system that integrates precision medicine and real-time prediction. This approach is opening a pathway towards the creation of foundational biological models able to achieve an appropriate balance between the amount of data utilized and the performance of the imposed biases \cite{Tierz2024}.

However, although the utility of physics-based hybrid learning for computational mechanics has been demonstrated \cite{Cai2021, Faroughi2024}, a bottleneck remains concerning graph networks, whose limitations arise from the computational complexity of the model. Large-scale simulations with fine meshes require substantial message passing to convey information across the entire domain \cite{veličković2022messagepassingway}, or a very fine time discretization, both of which significantly increase computational cost.

\section{Methodology}
\subsection{Problem statement}
\label{sec:prob_stat}

Let $\bs{{z}} \in \mathcal{M} \subseteq \mathbb{R}^n$ represent the independent state variables that fully characterize the dynamical system at a certain level of description. The set $\mathcal{M}$, known as the state space, contains all admissible states and is assumed to have the structure of a differentiable manifold in $\mathbb{R}^n$. The physical phenomenon under consideration can be modeled by a system of ordinary differential equations that govern the time evolution of each state variable $\bs{z}$ \cite{TIGNNs},
\begin{equation}\label{eq:PDE}
\dot{\bs{z}}= \dv{\bs{z}}{t} = F(\bs{z},t),\; t\in\mathcal{I}=(0,T],\; \bs{z}(0)= \bs{z}_0,
\end{equation}
where $t$ refers to the time coordinate in the time interval $\mathcal{I}$ and {$F(\bs{z}, t)$} refers to an arbitrary nonlinear function whose precise form is to be sought from data.

The main objective of this work is, given a time discretization consisting of \(N\) time steps, to determine the function \(F\) such that, for a given temporal state \(\bs{z}^t\), its temporal evolution can be calculated and, consequently, its next temporal state \(\bs{z}^{t+1}\) \cite{weinan2017proposal}. For this purpose, a temporal iterative scheme based on forward Euler integration is defined,
\begin{equation}\label{eq:EulerFoward}
\bs{z}^{t+1} = \bs{z}^t + \Delta t \cdot F(\bs{z}^t),
\end{equation}
wherein the function \( F(\bs{z}^t) \) is represented by a deep neural model \(\mathcal{N}\), which depends solely on the input \( \bs{z}^t \) and is parameterized by \( \bs \theta \) (the weights and biases of the network),
\begin{equation}\label{eq:Net}
F(\bs{z}^t) = \mathcal{N}(\bs{z}^t; \bs \theta).
\end{equation}

\subsection{Metriplectic structure: the GENERIC formalism}

To ensure that the resulting $F(\bs{z}^t)$-function conforms to the physical principles known in the scientific literature, it will be imposed in this work that it must have a metriplectic structure.

This is guaranteed by enforcing that $F$ adheres to the so-called General Equation for Non-Equilibrium Reversible-Irreversible Coupling (GENERIC) formalism \cite{ottinger1997dynamics}. This framework extends the classical Hamiltonian approach to include dissipative systems, offering an appealing alternative for describing soft tissue dynamics while ensuring thermodynamic correctness by construction \cite{grmela1997dynamics}.

Equivalently to what is described in Eq. (\ref{eq:PDE}), the GENERIC framework models the system's temporal evolution, $\dot{\bs{z}}$, governed through a set of state variables. It ensures that the resulting dynamic evolution satisfies fundamental thermodynamic principles throughout the process. The GENERIC formalism is composed by two blocks,
\begin{equation}\label{eq: Gen1}
    \dot{\bs{z}} = \{\bs{z}, E\} + [\bs{z}, S],
\end{equation}
where $\{\cdot, \cdot\}$ denotes the classic Poisson bracket and $[\cdot, \cdot]$ represents the dissipative bracket. To facilitate practical use, these brackets are often reformulated using two linear operators:
\begin{equation}\label{eq: Gen2}
    \bs{L} : T^*\mathcal{M} \rightarrow T\mathcal{M}, \quad \bs{M} : T^*\mathcal{M} \rightarrow T\mathcal{M},
\end{equation}
where \(T^*\mathcal{M}\) and \(T\mathcal{M} \) denote the cotangent and tangent bundles of the state space \(\mathcal{M}\), respectively. The operator \(\bs{L} = \bs{L}(\bs{z})\) corresponds to the Poisson bracket and must be skew-symmetric. Similarly, the friction matrix \(\bs{M} = \bs{M}(\bs{z})\) captures the irreversible behavior of the system and is symmetric and positive semi-definite, ensuring a non-negative dissipation rate \cite{TIGNNs}.

Given these components and considering a system discretized into \(n_v\) particles, the model must adhere to the laws of thermodynamics. To enforce this, we assume a metriplectic evolution \cite{guha2007metriplectic, morrison1986paradigm}, whose precise form is to be found from data:
\begin{equation}\label{eq: Gen3}
    \dot{\bs{z}} = \bs{L}(\bs{z}) \frac{\partial E}{\partial \bs{z}} + \bs{M}(\bs{z}) \frac{\partial S}{\partial \bs{z}},    
\end{equation}
where \(\bs{z} \in \mathbb{R}^{n_v \times n_{\text{dof}}}\) and \(n_{\text{dof}}\) represents the number of degrees of freedom of each particle.

The degeneracy conditions, 
\begin{equation}\label{eq: Gen101}
    \bs{L}(\bs{z}) \frac{\partial S}{\partial \bs{z}} = \bs{0}, 
\end{equation}
and

\begin{equation}\label{eq: Gen111}
    \bs{M}(\bs{z}) \frac{\partial E}{\partial \bs{z}} = \bs{0}, 
\end{equation}
along with the non-negativity of the irreversible bracket, ensure the adherence to the first (energy conservation) and second (entropy inequality) laws of thermodynamics. These can be expressed mathematically as
\begin{equation}\label{eq: Gen4}
    \frac{dE}{dt} = \{E, E\} = {0},
\end{equation}
and 
\begin{equation}\label{eq: Gen5}
    \frac{dS}{dt} = [S, S] \geq {0} .
\end{equation}
For a detailed proof of this assertion, the interested reader is directed to, for instance, \cite{grmela1997dynamics}, among many other possible references in the field.

\subsubsection{Nodal GENERIC implementation with a port-metriplectic formalism}

Thermodynamics assumes that a closed system does not exchange energy with the outside. In purity, this is not the case for a system such as the liver, which is subject to metabolic input from the body. However, at the time scale considered in this work (on the order of a certain number of seconds), the hypothesis that the liver is a closed system seems reasonable. The extension of the GENERIC formalism to open systems offers, however, no major conceptual difficulties, should it be deemed necessary \cite{PhysRevE.73.036126}.

Considering the liver as a closed system under a graph discretization approach poses, however, significant challenges. The main practical difficulty of this formalism lies in the assembly of the global matrices $\bs L$ and $\bs M$, whose size can be very large, depending on the number of particles in the discretisation. This breaks the local character of the equations, involving a considerable expenditure of memory and computational time in the computation and subsequent storage of these matrices.

In this paper we consider a reformulation of the GENERIC formalism in which each of the nodes forming the discretisation of the system (the liver) is considered as an open system that exchanges energy with its neighbours. Therefore, a reformulation of GENERIC will be done through the port-Hamiltonian framework for open systems \cite{Hern_ndez_2023Port, Tierz2024b}.

In this framework, each node symbolizes a portion of the bulk material, facilitating energy exchange with its neighboring nodes through designated ports, which correspond to the edges of the graph. In this context, the Poisson and dissipative brackets can be expressed as follows:
\begin{equation}\label{eq: Gen6}
    \{ \cdot, \cdot \} = \{ \cdot, \cdot \}_{\text{bulk}} + \{ \cdot, \cdot \}_{\text{bound}},
\end{equation}
and,
\begin{equation}\label{eq: Gen7}
    [ \cdot, \cdot ] = [ \cdot, \cdot ]_{\text{bulk}} + [ \cdot, \cdot ]_{\text{bound}}.
\end{equation}
In other words, both brackets are decomposed additively into bulk and boundary contributions. With this decomposition in mind, the GENERIC principle, Eq. (\ref{eq: Gen1}) applies to the bulk portion of the system associated to each node, that now is rewritten as
\begin{equation}\label{eq: Gen8}
    \dot{\bs{z}} = \{ \bs{z}, E \}_{\text{bulk}} + [ \bs{z}, S ]_{\text{bulk}} = \{ \bs{z}, E \} + [ \bs{z}, S ] - \{ \bs{z}, E \}_{\text{bound}} - [ \bs{z}, S ]_{\text{bound}}. 
\end{equation}

In matrix form, we obtain
\begin{equation}\label{eq: Gen9}
    \dot{\bs{z}} = \bs{L} \frac{\partial E}{\partial \bs{z}} + \bs{M} \frac{\partial S}{\partial \bs{z}} - \bs{L}_{\text{bound}} \frac{\partial E_{\text{bound}}}{\partial \bs{z}} - \bs{M}_{\text{bound}} \frac{\partial S_{\text{bound}}}{\partial \bs{z}}.
\end{equation}

In this case, the degeneracy conditions hold at the bulk level only as,
\begin{equation}\label{eq: Gen10}
    \bs{L}_{\text{bulk}}(\bs{z}) \frac{\partial S_{\text{bulk}}}{\partial \bs{z}} = \bs{0}, 
\end{equation}
and
\begin{equation}\label{eq: Gen11}
    \bs{M}_{\text{bulk}}(\bs{z}) \frac{\partial E_{\text{bulk}}}{\partial \bs{z}} = \bs{0}. 
\end{equation}

So if we plan to impose a GENERIC-like structure at a particle level, we must assume that this particle receives energy input/outputs from surrounding particles, as dictated by the graph structure \cite{Tierz2024b}.

Assume that a given particle \( i \) with state variables \( \bs{z}_i \in \mathbb{R}^{n_{\text{dof}}} \) is connected in the graph with \( j = 1, \ldots, n_{\text{neigh}} \) neighbouring particles. Its state variables will thus evolve in time as
\begin{equation}\label{eq: Gen12}
\dot{\bs{z}}_i = \bs{L}_i(\bs{z}_i) \frac{\partial e_i}{\partial \bs{z}_i} + \bs{M}_i(\bs{z}_i) \frac{\partial s_i}{\partial \bs{z}_i} - \sum_{j=1}^{n_{\text{neigh}}} \left( \bs{L}_{ij}(\bs{z}_j) \frac{\partial e_j}{\partial \bs{z}_j} + \bs{M}_{ij}(\bs{z}_j) \frac{\partial s_j}{\partial \bs{z}_j} \right). 
\end{equation}

The total energy and entropy of the system are obtained by assuming
\begin{equation}\label{eq: Gen13}
E = \sum_{i=1}^{n_v} e_i,
\end{equation}
being $e_i$ the energy contribution per node, an
\begin{equation}\label{eq: Gen14}
S = \sum_{i=1}^{n_v} s_i.
\end{equation}
where $s_i$ is equivalently the entropy contribution per node. The degeneracy conditions at the bulk level play here the role of the particle level, so that
\begin{equation}\label{eq: Gen15}
\bs{\bs{L}}_i(\bs{z}_i) \frac{\partial s_i}{\partial \bs{z}_i} = \bs{0},
\end{equation}
and
\begin{equation}\label{eq: Gen16}
\bs{M}_i(\bs{z}_i) \frac{\partial e_i}{\partial \bs{z}_i} = \bs{0},
\end{equation}
for all \( i = 1, \ldots, n_v \). Defining as a whole the imposition of the metriplectic structure for the hybrid framework, schematically shown in the Figure \ref{fig:hybridnet}.

This modification of the GENERIC formalism ensures compliance with the equations of thermodynamics and, at the same time, avoids the assembly of the global matrices, leaving the problem as the determination, from the available data, of the precise shape of the GENERIC structure at the scale of each model node.

\subsection{Geometric structure: Graph Neural Networks}

Apart from respecting the dictates of the laws of thermodynamics, the other major ingredient of this work is the desire to develop a technique that allows for the particularisation of predictions for anatomies not previously seen in training. This task is addressed by using the geometric biases present in graph neural networks \cite{bronstein2017geometric}.

\begin{figure}[h]
    \centering
    \includegraphics[scale=0.9]{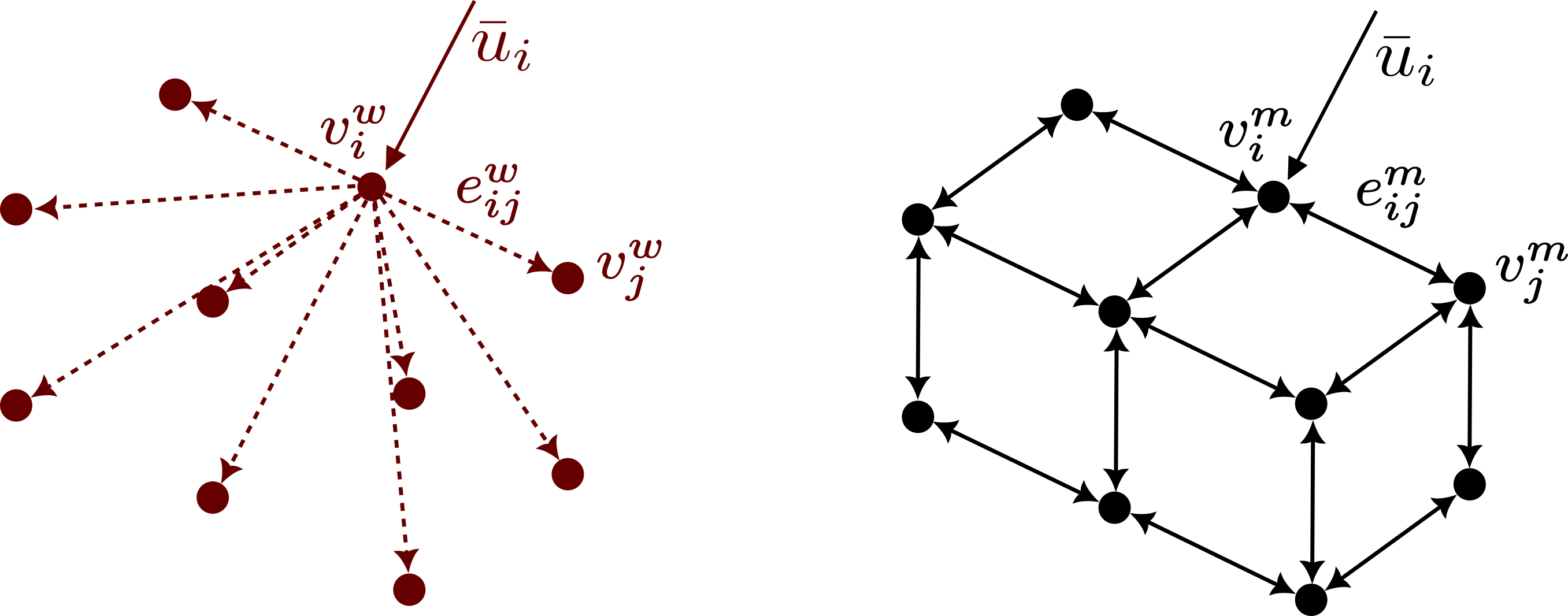}
    \caption{Graph representations of the unidirected actuator graph, \( \mathcal{G}^w \) (in red) and the bidirected central/mesh graph, \( \mathcal{G}^m \) (in black). In each graph, $\bs{v}_i$ represents node attributes, $\bs{e}_{ij}$ denotes the edge attributes, $\bar{\bs u}_i$ corresponds to external displacements or interactions. Thus, the complete representation of this example would be defined as the superposition of both graphs, forming the multi-graph system $G$.}
    \label{fig:graphs}
\end{figure}

Let \( \mathcal{G} = (\mathcal{V}, \mathcal{E}, \bs{u}) \) be a directed graph, where \( \mathcal{V} = \{1, \dots, n\} \) is a set of \( |\mathcal{V}| = n_v \) vertices, \( \mathcal{E} \subseteq \mathcal{V} \times \mathcal{V} \) is a set of \( |\mathcal{E}| = e \) edges and \( \bs{u} \in \mathbb{R}^{F_g} \), a feature vector, where $F_g$ denotes the dimension of this global feature vector, containing global parameters of the simulation (such as density, non-dimensional parameters, etc.). Each vertex and edge in the graph is associated with a node and a pairwise interaction between nodes respectively in a discretized physical system. The global feature vector defines the properties shared by all the nodes in a graph, such as gravity or material properties. For each vertex \( i \in \mathcal{V} \) we associate a feature vector \( \bs{v}_i \in \mathbb{R}^{F_v} \), where $F_v$ represents the dimensionality of the vector embedded within the node, capturing the physical properties associated with each individual node. Similarly, for each edge \( (i, j) \in \mathcal{E} \) we associate an edge feature vector \( \bs{e}_{ij} \in \mathbb{R}^{F_e} \) \cite{TIGNNs}. In this framework, the collection of graphs \( \{ \mathcal{G}_1, \dots, \mathcal{G}_M \} \) together builds the overall multi-graph \( G_k,\;  \text{with } k = 1,\ldots, M  \), where each individual graph is \( \mathcal{G}_M \). Figure \ref{fig:graphs} illustrates a multi-graph example of a simple mesh composed of two hexahedral elements. The spatial discretization forms the central graph, shown in black, while an actuator graph, shown in red, is originated at the loaded node. Together, these graphs provide a more detailed and richer representation of the system.

The implementation of multi-graphs in this context allows for complementing the central graph \( \mathcal{G}^m \), which represents the finite element mesh, by adding new subgraphs \( \mathcal{G}^w \), generated from the loaded node. These new subgraphs, connected to the entire domain $\Omega$, enable the transmission of message flows, starting from the loaded position.

For the central graph , the edges \( \bs{e}^m_{ij} \) collect the information corresponding to the relative deformation of the model. This information will include both the relative distance during the input time step, denoted as \( \bs{q}_{ij}(\bs{z}^t) \) and the relative distance in the initial unloaded state \( \bs{q}_{ij}(\bs{z}^0) \). Similarly, the nodes \( \bs{v}^m_i \) will store an embedding vector with velocity \( \bs{v}_i(\bs{z}^t) \), stress state \( \bs \sigma(\bs{z}^t) \), and flags \( \bs{n}_i \) indicating whether a node is fixed, loaded, or in a neutral state by default.

In the actuator graphs, equivalent information will be encoded in the edges \( \bs{e}^w_{ij} \); however, the vertices \( \bs{v}^w_i \) will contain data regarding the imposed displacement \( \bar{\bs u}_i \).

\subsubsection{Encoding}

For the initial encoding structure \( \epsilon \) we use four MLPs, \(\epsilon = \{ \epsilon^m_e, \epsilon^w_e, \epsilon^m_v, \epsilon^w_v \}\), to transform the vertex and edge initial feature vectors into the latent space, two for the central graph attributes, $\bs{x}^w_i \in \mathbb{R}^{F_l}$ and $\bs{x}^w_{ij} \in \mathbb{R}^{F_l}$ and two for the actuators graphs respectively,
$\bs{x}^m_i \in \mathbb{R}^{F_l}$ and $\bs{x}^m_{ij} \in \mathbb{R}^{F_l}$.

\begin{equation}\label{eq:EncoderE}
    \epsilon_e : \mathbb{R}^{F_e} \longrightarrow \mathbb{R}^{F_l}, \quad \bs{e}_{ij} \longmapsto \bs{x}_{ij}, \quad \bs{X} = \{ \bs{x}_{ij} \mid (i, j) \in \mathcal{E} \}
\end{equation}

\begin{equation}\label{eq:EncoderV}
    \epsilon_v : \mathbb{R}^{F_v} \longrightarrow \mathbb{R}^{F_l}, \quad \bs{v}_i \longmapsto \bs{x}_i, \quad \mathcal{X} = \{ \bs{x}_{i} \mid i \in \mathcal{V} \}
\end{equation}

The latent space dimension $F_l$ is set as a hyperparameter to regulate its expressiveness, as well as way to control the number of trainable parameters that define the algorithm.

\subsubsection{Processor}

The processor plays a central role in the algorithm, as it facilitates the exchange of nodal information between vertices through message passing, transmitting the flow of information between the actuators and the central mesh through the latent space. When subdividing each step of the processor, two distinct stages are defined. In the first stage, information from the nodes is gathered in conjunction with the edges embeddings, $\bs x_{ij}$ and integrated into a MLP, applied to both the edges of the central graph, $f^{m}$ and those of the actuator graph, $f^{w}$. In the second stage, the latent representation of the edges is incorporated through a permutation-invariant aggregation function, $\phi$ , combining it with the node embedding in another MLP, $f^v$ to obtain the latent representation of the vertices. This process is repeated for a number of hyperparameterized steps, adding the latent representations in a residual manner during the loop \cite{pfaff2021learningmeshbasedsimulationgraph},
\begin{equation}\label{eq:Processor}
    f^{m}(\bs{x}^{m}_{ij}, \bs{x}^m_i, \bs{x}^m_j) \rightarrow \bs{x}'^{m}_{ij}, \quad f^{w}(\bs{x}^{w}_{ij}, \bs{x}^w_i, \bs{x}^w_j) \rightarrow \bs{x}'^{w}_{ij}, \quad f^{v} \left(\bs{x}_i, \phi (\bs{x}'^{m}_{ij}), \phi (\bs{x}'^{w}_{ij}) \right) \rightarrow \bs{x}'_{i}
\end{equation}
where $(\cdot, \cdot, \cdot)$ denotes vector concatenation and $\bs x'_i$ and $\bs x'_{ij}$ are the updated nodal and edge latent representations.

\subsubsection{Nodal vanilla decoder}

\begin{figure}[h]
    \hspace{-1.2cm}
    \includegraphics[scale=0.21]{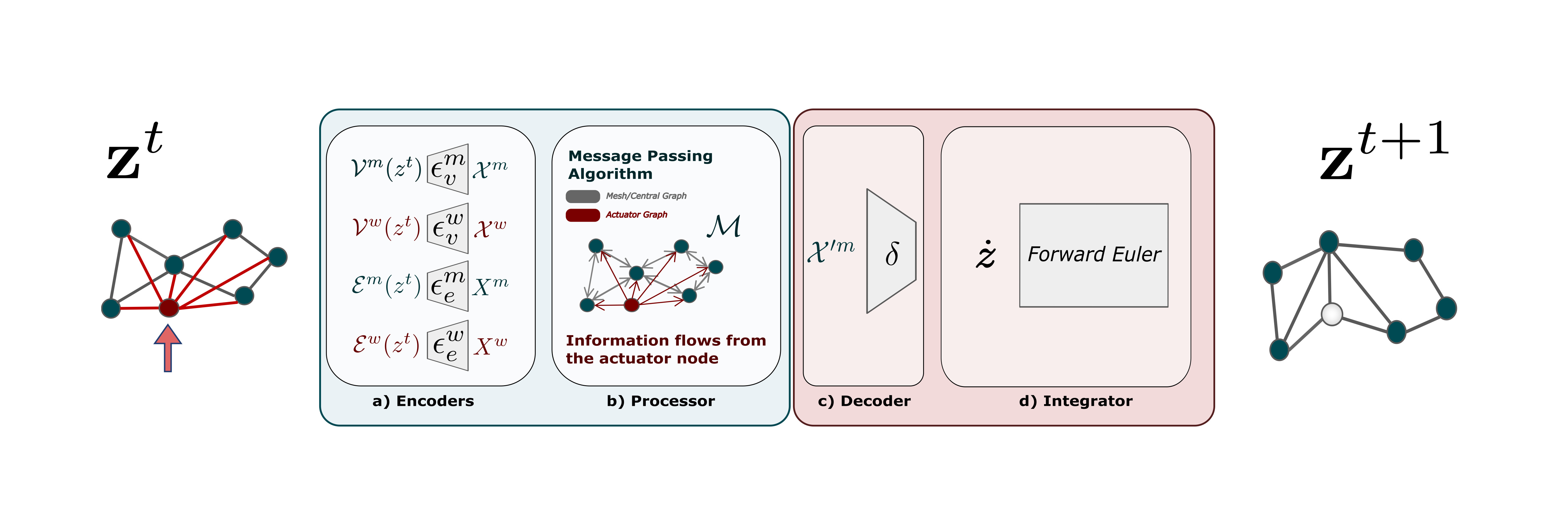}
    \caption{Overview of the algorithm block scheme for predicting single-step state variable changes in the vanilla graph framework. (a) The encoder transforms the node and edge features into the latent space. (b) The processor distributes information across the graph using  $\mathcal{M}$  message-passing modules. (c) The decoder extracts the gradient vectors from the processed graph. (d) The integrator predicts the state variables for the next time step. This entire sequence is repeated iteratively to generate the system’s dynamic rollout.}
    \label{fig:dd}
\end{figure}

For the vanilla data-driven framework the last block extracts the total derivative, $\dot{\bs{z}}$, directly from the latent vectors of the mesh nodes with a single decoder $\delta$, therefore defining the temporal evolution of the model \( \dot{\bs{z}}_i \in \mathbb{R}^{F_{\bs{z}}} \), where $F_z$ denotes the dimension of the state variables, as seen in Fig. \ref{fig:dd}, 
\begin{equation}
    \delta : \mathbb{R}^{F_l} \longrightarrow \mathbb{R}^{F_{\bs{z}}}, \quad \bs{x}'^m_i \longmapsto \dot{\bs{z}}_i, \quad \mathcal{X'} = \{ \bs{x}'_{i} \mid i \in \mathcal{V} \}   .
\end{equation}

\subsubsection{Nodal physics-informed decoder}

For the hybrid framework the last block extracts as well the relevant physical output information of the system from the node latent feature vector \( \bs{x}'_i \in \mathbb{R}^{F_{y,v}} \) and from the edge latent feature vector \( \bs{x}'_{ij} \in \mathbb{R}^{F_{y,e}} \), implemented with two decoders built by independent MLPs (\( \delta_v, \delta_e \)),
\begin{equation}\label{eq: Decoder1}
    \delta_v : \mathbb{R}^{F_l} \longrightarrow \mathbb{R}^{F_{y,v}}, \quad \bs{x}'^m_i \longmapsto \left( \frac{\partial E}{\partial \bs{z}}, \frac{\partial S}{\partial \bs{z}} \right),
\end{equation}
\begin{equation}\label{eq: Decoder2}
    \delta_e : \mathbb{R}^{F_l} \longrightarrow \mathbb{R}^{F_{y,e}}, \quad \bs{x}'^m_{ij} \longmapsto (\bs{l}, \bs{m}).
\end{equation}
In each case, the subindex \( v \) corresponds to the nodal decoding of the energy gradient \( \frac{\partial \bs{E}}{\partial \bs{z}} \) and the entropy gradient \( \frac{\partial \bs{S}}{\partial \bs{z}} \). The subindex \( e \), on the other hand, refers to the decoding process of the flattened operators \( \bs{l} \) and \( \bs{m} \) from each edge, also considering the self-loop.

\begin{figure}[h]
    \centering
    \includegraphics[scale=0.19]{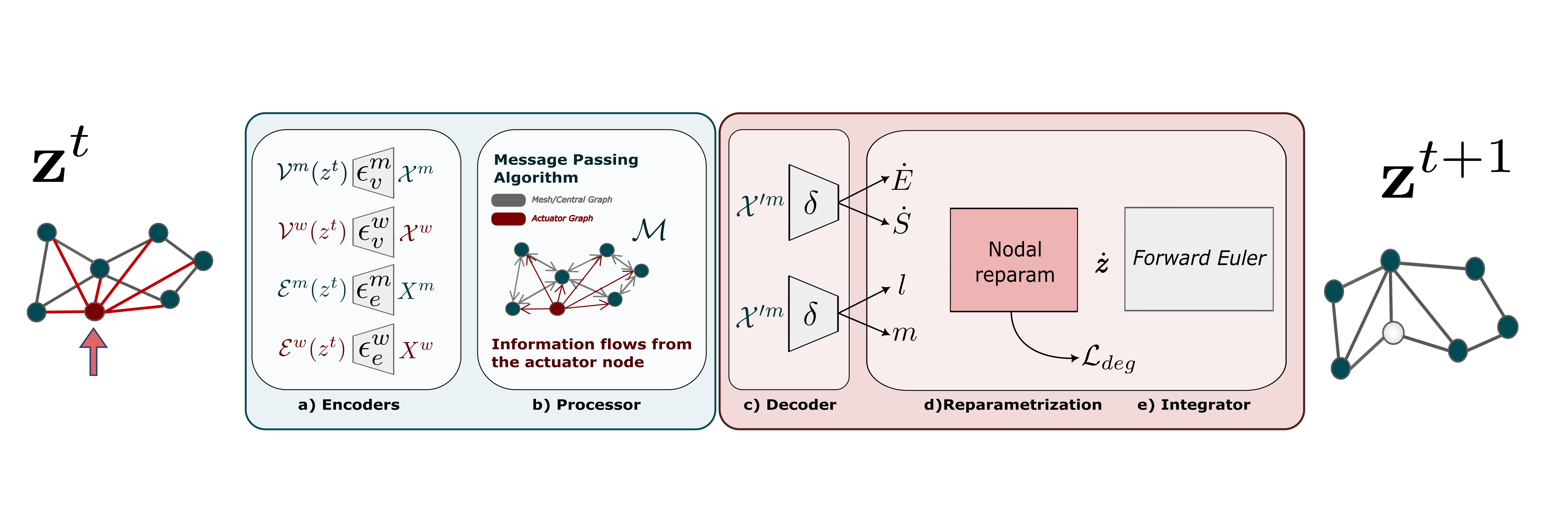}
    \caption{Overview of the algorithm block scheme for predicting single-step state variable changes in the hybrid graph framework. (a) The encoder transforms the node and edge features into the latent space. (b) The processor distributes information across the graph using  $\mathcal{M}$  message-passing modules. (c) The decoder extracts the energy and entropy gradients from each node in addition to the  flattened operators from each edge. (d) The reparametrization agregates the boundary terms as seen in Eq. (\ref{eq: Gen12}) to each node subsystem, predicting a gradient vector based on the GENERIC formulation (e) The integrator predicts the state variables for the next time step. This entire sequence is repeated iteratively to generate the system’s dynamic rollout.}
    \label{fig:hybridnet}
\end{figure}

The extraction of all these components, instead of directly extracting the gradients, has a primary purpose: the imposition of a metriplectic structure that restricts the network to solutions that geometrically satisfy the thermodynamics of the system, with the added objective of imparting a physical bias to the network to enhance its robustness in purely data-driven scenarios. Fig. \ref{fig:hybridnet} illustrates this implementation through the metriplectic imposition, incorporating hard constraints for constructing linear operators alongside soft constraints applied to the loss function.

\subsection{Learning procedure and data properties}

\begin{figure}[h!]
\centering
\includegraphics[width=\textwidth, trim= 100 120 100 120,clip]{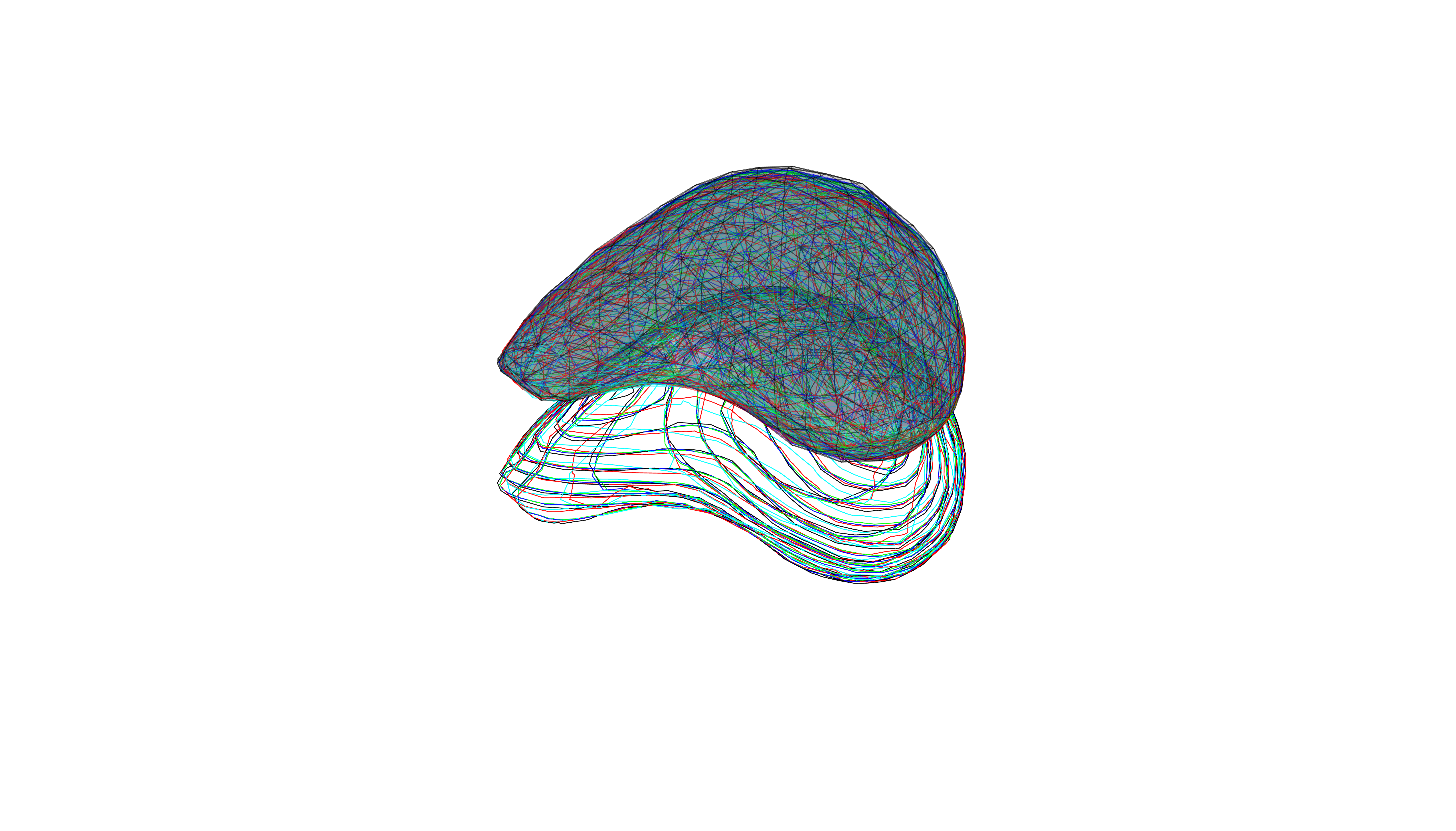} 
\captionsetup{skip=6pt} 
\caption{Overlay of the different hepatic geometries used, along with a projection of the representative contour curves. Every color is associated with a different anatomy, providing a visual representation of the anatomical variability inherent to the database. Image top - meshes. Image bottom - level set curves for each anatomy.}
\label{fig:liver_plots}
\end{figure}

The master database contains 190 simulations across five distinct anatomies \cite{li2023medshapenet}, coined as $\{\bs{L1},\ldots,\bs{L5}\}$. They have mesh sizes ranging from 680 to 460 per mesh. We assume Dirichlet boundary conditions on the visceral face (Fig. \ref{fig:liver_plots} and Table \ref{tab:liver_mesh_data}). Data normalization involves standardizing the state tensor, which comprises a 12-element vector per node for each snapshot and simulation. This vector includes three positional dimensions, three velocity dimensions, and six components of the Cauchy stress tensor.

\begin{table}[h!]
\centering
\begin{tabular}{|c|c|c|c|}
\hline
{Geometry} & {Volume [$m^3$]} & {Nodes} &{Elements} \\ \hline
$\bs{L1}$ & 2.71E-4 & 530  & 1734 \\ \hline
$\bs{L2}$ & 2.33E-4 & 460  & 1449 \\ \hline
$\bs{L3}$ & 2.14E-4 & 492  & 1547 \\ \hline
$\bs{L4}$ & 2.58E-4 & 680  & 2299 \\ \hline
$\bs{L5}$ & 2.61E-4 & 665  & 2203 \\ \hline
\end{tabular} \vspace{0.4cm}
\caption{Specific data for each anatomy and, consequently, for the meshes resulting from its spatial discretization. The training subset covers from $\bs{L1}$ to $\bs{L4}$, while $\bs{L5}$ is reserved for testing on a geometry unknown to the network.}
\label{tab:liver_mesh_data}
\end{table}

Of the five geometries, four are used for training, resulting in 760 simulations, each with 20 time steps. Each simulation applies nodal displacements to a selection of 1 to 3 nodes, with traction or compression varying between 0.5 cm and 2.5 cm. For validation, 20\% of the simulations from the remaining geometry ($\bs{L5}$) are used. The primary testing dataset is based on the 190 total simulations of $\bs{L5}$, including those also used for validation to avoid biases.

The frameworks---both vanilla and hybrid---are designed with approximately 240,000 trainable parameters. The training process utilizes the Adam optimizer along with a learning rate scheduler that reduces the rate once a plateau in training loss is reached. Model evaluation includes five standard models and one fast inference model.

Training takes approximately two days on an RTX 3090, while weight selection for test inference is done through early stopping based on the mean relative rollout error across the complete test dataset.

Both methodologies are formulated as minimization problems for a given loss function, $\mathcal{L}$. In the case of the vanilla model $\mathcal{L}_{\text{vanilla}}$, the cost function only evaluates the data loss $\mathcal{L}_{\text{data}}$ representing as the MSE along the graph nodes and state variables between the predicted and the ground-truth time derivative of the state vector in a given batch of snapshots,
\begin{equation}\label{eq:lossdata}
    \mathcal{L}_{\text{vanilla}} = \mathcal{L}_{\text{data}} = \frac{1}{N} \sum_{i=0}^{N} \frac{1}{D} \left\| \frac{d \bs{z}_i^{\text{GT}}}{dt} - \frac{d \bs{z}_i^{\text{net}}}{dt} \right\|_2^2,
\end{equation}
being $N$ the total number of nodes per batch and $D$ the total number of features.

The hybrid model loss $\mathcal{L}_{\text{hybrid}}$, in contrast, incorporates additional loss terms that account the residual of the degeneracy conditions, Eqs. (\ref{eq: Gen15}) and (\ref{eq: Gen16}),  
\begin{equation}\label{eq:lossdeg}
    \mathcal{L}_{\text{deg}} = \frac{1}{N} \sum_{i=0}^{N} \left\| \bs{L} \frac{\partial S}{\partial \bs{z}_i} + \bs{M} \frac{\partial E}{\partial \bs{z}_i} \right\|_2^2,
\end{equation}
modulated by the parameter $\lambda_d$ to control the influence of the physical term,
\begin{equation}\label{eq: losshib}
    \mathcal{L}_{\text{hybrid}} = \mathcal{L}_{\text{data}} + \lambda_d  \mathcal{L}_{\text{deg}}.
\end{equation}
The value of $\lambda_d$ is set to 5 for all trials, as sensitivity analyses have determined that its optimal range lies between 1 and 20,

\subsubsection{Material properties}

To determine the $F(\bs{z}^t)$-function, synthetic data will be employed. These will be obtained by high-fidelity finite element simulations assuming a homogeneous simplification of liver tissue, omitting vascularization and other relevant anatomical features. The hyperelastic part of the response response will be defined by a strain energy density function  $W$.

To model this hyperelastic behavior, the Ogden model \cite{ogden1997nonlinear} will be used,
\begin{equation}
W = \sum_{i=1}^{N} \frac{\mu_i }{\alpha_i}(\lambda_{1}^{\alpha_i} + \lambda_{2}^{\alpha_i} + \lambda_{3}^{\alpha_i} - 3) + \sum_{i=1}^{N} \frac{1}{d_i}(J - 1)^{2i},
\label{eq:Ogden}
\end{equation}
by assuming $N=3$, where:
\begin{itemize}
    \item \( W \): the strain energy density function for the hyperelastic material.
    \item \( \mu_i \): material parameters that define the contribution of each mode to the energy density.
    \item \( \alpha_i \): exponents that characterize the non-linear behavior of the material.
    \item \( \lambda_1, \lambda_2, \lambda_3 \): the principal stretches of the material.
    \item \( J \): the determinant of the deformation gradient, representing the volumetric change of the material.
    \item \( d_i \): parameters related to the volumetric response, influencing the contribution of the volumetric term.
    \item \( N \): the number of terms in the summation, allowing for a more detailed representation of the material’s response.
\end{itemize}

Alongside, the viscous part of the behaviour will be modeled through Prony series, which are suitable for capturing the time-dependent behavior of the material \cite{park1999methods}. The Prony series representation is given by
\begin{equation}
G(t) = G_0 \left(\alpha_{\infty} + \sum_{k=1}^{N} \alpha_k e^{-t/\tau_k}\right),
\label{eq:Prony}
\end{equation}
where:
\begin{itemize}
    \item \( G(t) \): the dynamic shear modulus as a function of time in pascals.
    \item \( G_0 \): the initial shear modulus (the value at \( t = 0 \)) in pascals.
    \item \( \alpha_{\infty} \): represents the long-term behavior of the material (the steady-state value).
    \item \( \alpha_k \): coefficients that weight the contribution of each relaxation mode.
    \item \( \tau_k \): relaxation times corresponding to each mode, dictating how quickly the material responds over time.
    \item \( N \): the number of terms used in the series, allowing for a more accurate representation of the material's behavior.
\end{itemize}

\begin{figure}[h!]
    \centering
    \includegraphics[width=\linewidth]{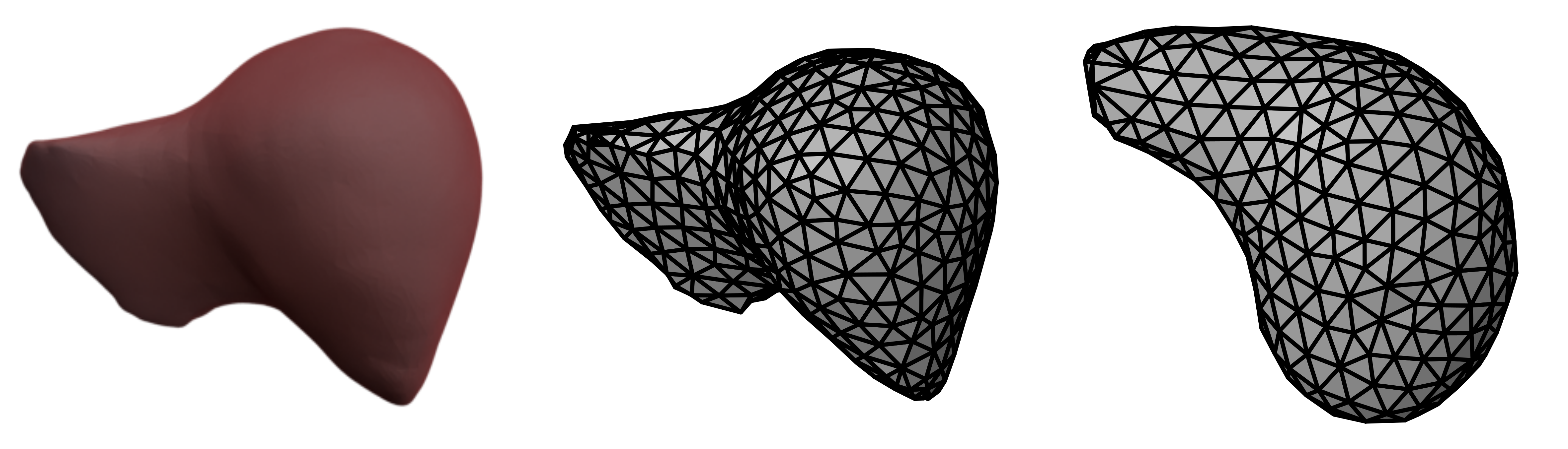}
    \caption{Finite element discretization of the liver. This figure demonstrates the meshing strategy designed to accurately capture the geometric and structural intricacies of hepatic tissue, providing a reliable foundation for computational simulations. The mesh is composed of 4-node linear tetrahedral elements. The illustrations show: the rendering of the liver parenchyma (left), the corresponding mesh for this perspective (center), and the equivalent mesh representation of the liver's anterior face (right).}
    \label{fig:liver1}
\end{figure}

Finally, the material parameters will be calibrated based on existing experimental literature to ensure the model accurately reflects the properties of hepatic tissue, with parameter values detailed in Table \ref{tab:ogden_prony} \cite{martinez2014}. For precise simulations, the liver anatomy is discretized using finite elements, as shown in Figure \ref{fig:liver1}, which highlights the meshing strategy and the geometric detail captured to represent the hepatic structure.

\begin{table}[h!]
    \centering
    \begin{tabular}{|c|c|c|}
    \hline
    \textbf{Model} & \textbf{Parameter} & \textbf{Value} \\ \hline
    \multirow{3}{*}{Ogden} 
        & $\text{E}$ (Pa) & 12,598 \\ \cline{2-3}
        & $\mu_L$ (Pa) & 364.74 \\ \cline{2-3} 
        & $\alpha_L$ & 16.19 \\ \cline{2-3} 
        & $d$ (Pa$^{-1}$) & $9.525 \times 10^{-5}$ \\ \hline
    \multirow{4}{*}{Prony Series} 
        & $\alpha_1$ & 0.56057 \\ \cline{2-3} 
        & $\tau_1$ (s) & 6.36173 \\ \cline{2-3} 
        & $\alpha_2$ & 0.40585 \\ \cline{2-3} 
        & $\tau_2$ (s) & 0.73348 \\ \hline
    \end{tabular}
    \vspace{0.4cm}
    \caption{Parameters of the Ogden model and Prony series constants.}
    \label{tab:ogden_prony}
\end{table}

\section{Results}
\label{sec:results}

In this section, we present and compare the results of the two approaches considered: the vanilla graph network, assumed as the baseline, and the hybrid graph neural network incorporating the metriplectic structure. To ensure a fair comparison, both methods are tuned to have an equal number of parameters. Additionally, each model is trained using different seeds to ensure reproducibility. All evaluations are also conducted on a dataset of 190 simulations derived from an unseen anatomy, referred to as \textit{Test} from this point forward.

\subsection{Evaluation metrics}

The Root Mean Square Error (RMSE) values 
\begin{equation}\label{eq:rmse}
    \text{RMSE}(\bs z_{\text{ref}}, \bs z_{\text{pred}}) = \sqrt{\frac{1}{n_{\text{snp}}} \sum_{i=1}^{n_{\text{snp}}} \left( \frac{1}{n_v} \sum_{j=1}^{n_v} \frac{\|  \bs z_{\text{ref}} -  \bs z_{\text{pred}} \|_2^2}{D} \right) },
\end{equation}
in Table \ref{tab:rmse_5p} show that the hybrid model outperforms the vanilla model across all considered state variables. Specifically, the hybrid approach achieves lower RMSE values for \(\bs{q}\), \(\bs{v}\), and \(\bs{\sigma}\), suggesting that the incorporation of the metriplectic structure improves the model’s predictive accuracy.

Let \( n_v \) represent the total number of nodes per mesh, and \( n_{\text{snp}} \) the number of snapshots taken per simulation. In this case, all snapshots from each \textit{Test} subset are considered collectively.

\begin{table}[h]
    \centering
    \begin{tabular}{@{}lcc@{}}
        \toprule
        & \textbf{Hybrid} & \textbf{Vanilla} \\ \midrule
        $\bs{q}$ (m): & $(3.730 \pm 0.06) \times 10^{-4}$ & $(4.060 \pm 0.11) \times 10^{-4}$ \\
        $\bs{v}$ (m/s): & $(7.430 \pm 0.06) \times 10^{-4}$ & $(8.430 \pm 0.34) \times 10^{-4}$ \\
        $\bs{\sigma}$ (Pa): & $(1.664 \pm 0.009) \times 10^{3}$ & $(1.755 \pm 0.037) \times 10^{3}$ \\ 
        \bottomrule
    \end{tabular}
    \vspace{0.4cm}
    \caption{RMSE comparison between three Hybrid and Vanilla Models across all snapshots in the testing subset, evaluated over five passes and approximately 200,000 trainable parameters. The \textit{Test} subset includes 190 untrained simulations in an entirely new anatomy, coined as $\bs{L5}$.}
    \label{tab:rmse_5p}
\end{table}

To further assess the models' performance, we compute the Relative Root Mean Square Error (RRMSE), 
\begin{equation}\label{eq:rrmse}
    \text{RRMSE}(\bs z_{\text{ref}}, \bs z_{\text{pred}}) = \sqrt{ \frac{1}{n_{\text{snp}}} \sum_{i=1}^{n_{\text{snp}}} \left( \frac{1}{n_v} \sum_{j=1}^{n_v} \frac{\| \bs z_{\text{ref}} - \bs z_{\text{pred}} \|_2^2}{\| \bs z_{\text{ref}} \|_\infty^2} \right) },
\end{equation}
as shown in Table \ref{tab:rrmse_5p}. Similar to the RMSE results, the hybrid model demonstrates superior performance by consistently producing lower RRMSE values across all state variables.

\begin{table}[h]
    \centering
    \begin{tabular}{@{}lcc@{}}
        \toprule
        & \textbf{Hybrid} & \textbf{Vanilla} \\ \midrule
        $\bs{q}$: & $(1.2713 \pm 0.021) \times 10^{-3}$ & $(1.383 \pm 0.038) \times 10^{-3}$ \\
        $\bs{v}$: & $(5.18567 \pm 0.1038) \times 10^{-2}$ & $(5.9900 \pm 0.2299) \times 10^{-2}$ \\
        $\bs{\sigma}$: & $(2.73333 \pm 0.1066) \times 10^{-2}$ & $(2.8893 \pm 0.0624) \times 10^{-2}$ \\ 
        \bottomrule
    \end{tabular}
    \vspace{0.4cm}
    \caption{RRMSE comparison between three Hybrid and Vanilla Models across all snapshots in the testing subset, evaluated over five passes and approximately 200,000 trainable parameters. The \textit{Test} subset includes 190 untrained simulations in an entirely new anatomy, coined as $\bs{L5}$.}
    \label{tab:rrmse_5p}
\end{table}

To extend the discussion, other models with longer-range message passing procedures are explored in Appendix \ref{an:12p}.

\subsection{Inference performance analysis for each simulation rollout}

\begin{figure}[h]
 \includegraphics[width=\linewidth]{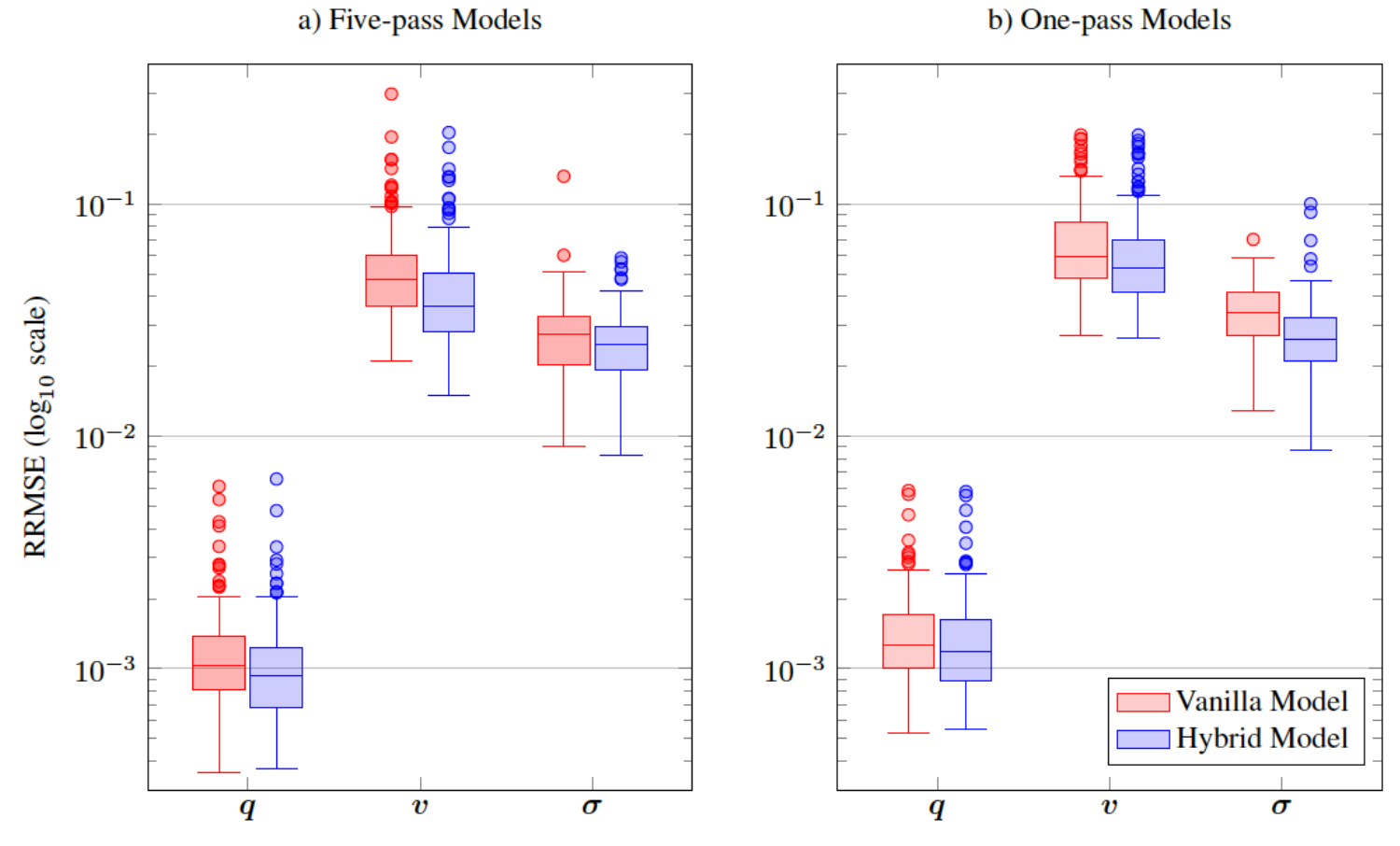}
    \caption{Comparison of RRMSE rollouts between the five-pass and one-pass models across 190 unseen simulations in a new geometry. The blue boxes represent the error distribution for the vanilla model, while the red boxes show the distribution for the hybrid model.} 
    \label{fig:boxplot_20k_combined} 
\end{figure}

Absolute and relative metrics are excellent indicators for gaining a general overview of the model's performance, however, it is essential not to lose sight of the fact that one of the main objectives is to ensure adequate robustness during the full temporal evolution, avoiding significant impacts from the residual error accumulations generated by the recurrent inference on the model. First, the boxplots presented in Fig.\ref{fig:boxplot_20k_combined} compare the distribution of relative error (RRMSE) for the three state variables in each simulation rollout.

These results correlate with the previously mentioned metrics, demonstrating that the hybrid methodology enhances both the predictive capacity and the robustness of the model, for architectures trained with five message passing steps as well as with one.

\begin{figure}[h]
    \centering
    \includegraphics[scale=0.8]{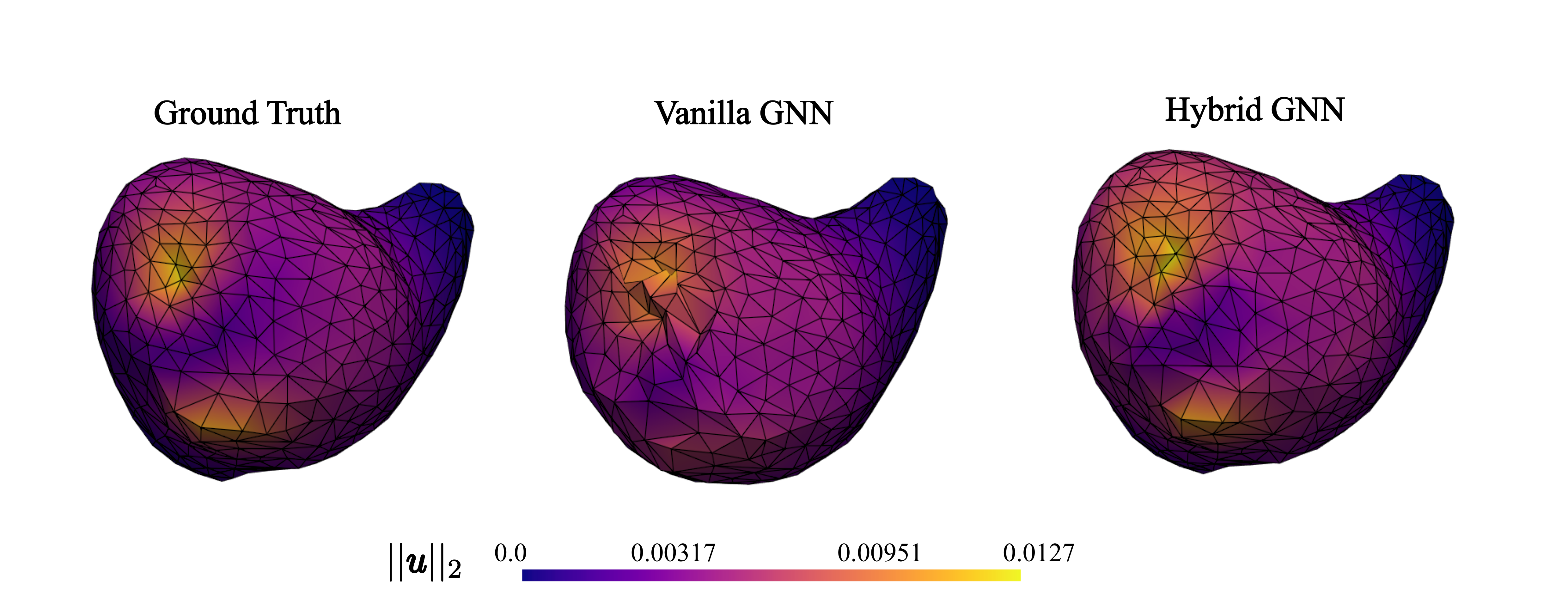}
    \caption{Comparison of the final inference step between the two proposed models and the finite element solution, used as ground truth, for previously unseen geometries and load conditions. The color scale represents the 2-norm of the displacement vector on each node.}
    \label{fig:comp1}
\end{figure}

\begin{figure}[h]
    \centering
    \includegraphics[scale=0.8]{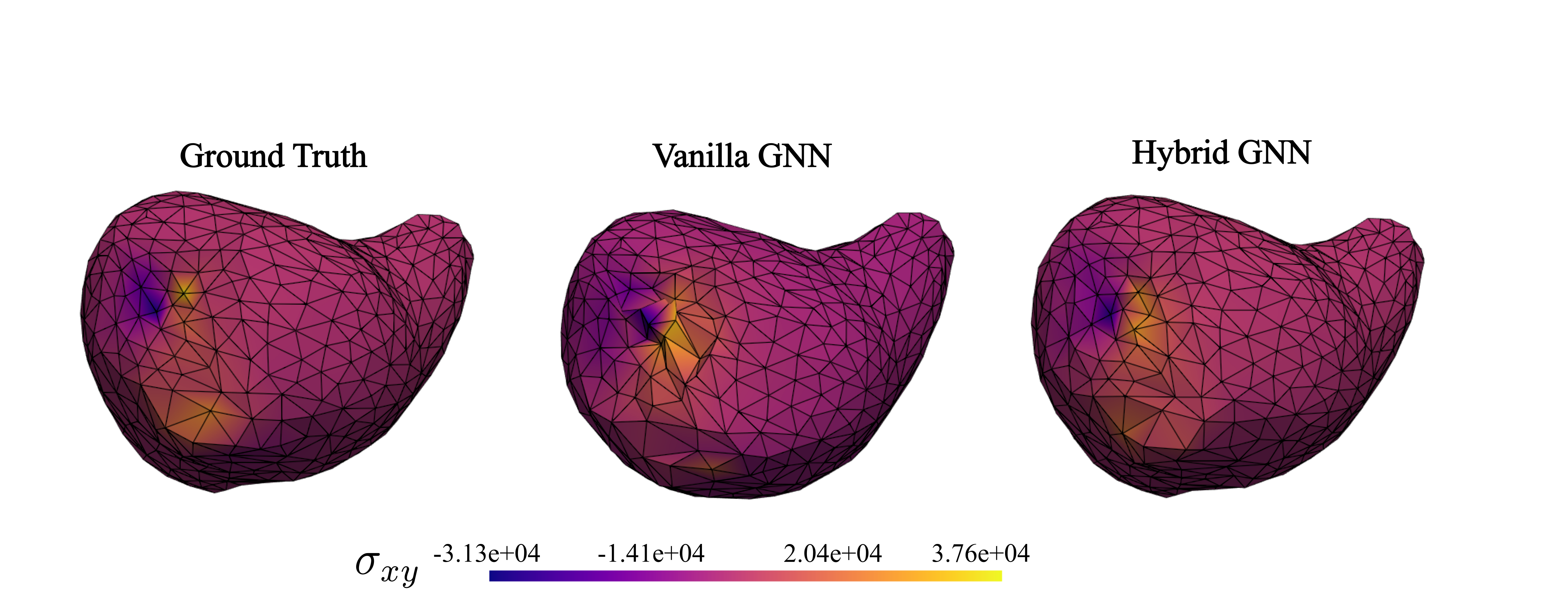}
    \caption{Comparison of the final inference step between the two proposed models and the numerical solution, used as ground truth, for untrained geometries and unseen load conditions. The color scale represents the $\sigma_{xy}$ scalar value on each node.}
    \label{fig:comp2}
\end{figure}

Although the difference can be appreciated in terms of relative or absolute error, it is important to note that this is measured at the full graph level, which makes the predictive improvement associated with the error reduction hard to comprehend using only global metrics. Figs. \ref{fig:comp1} and \ref{fig:comp2} support the main argument, illustrating the effect of physics bias in a simulation with an untrained geometry. 

The main hypothesis suggests that the metriplectic structure, which enforces both weak and strong constraints, improves the stability of nodal inference and prevents stress-induced error spikes from negatively impacting positional prediction. This aligns with the observation that significant error accumulation tends to occur around areas of imposed displacement and near stress concentrators.

The maximum response times exceed 600 Hz for one-step models, while the average times for the more refined models hover around 150 Hz for both methodologies. However, a more detailed description of the evolution of the response time with an increased pass count in the graph processor can be found in the Appendix \ref{an:time}.

\subsection{Robustness against previously unseen anatomies}

Another argument that complements the high feedback rates of the network is its robustness against untrained anatomies and meshes, making it highly suitable for precision medicine contexts using digital twins. To further evaluate this capability, we have supplemented the main test database with 33 new and untrained simulations within an already trained geometry, designating this subset as \textit{Extra}. This subset serves as an intermediate point between the 190 \textit{Train} sampled simulations and the 190 \textit{Test} simulations, which are used to assess the rollout performance.

\begin{figure}[h]
\centering  \includegraphics[width=0.8\linewidth]{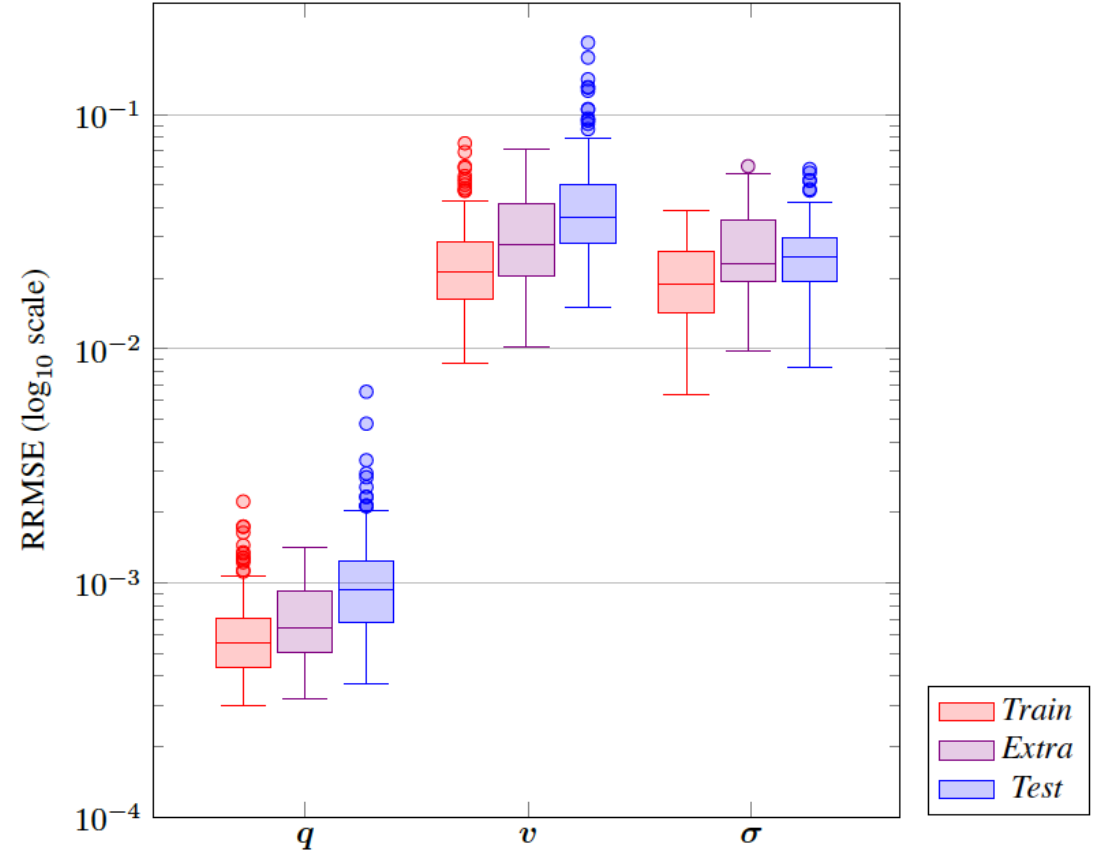}
     \caption{Comparison of rollout Relative Root Mean Square Error (RRMSE) across three datasets: \textit{Train} (trained data), \textit{Extra} (unseen load states with known anatomy), and \textit{Test} (untrained anatomies and load states, main testing dataset) for the parameters \(\bs q\), \(\bs v\), and \(\bs \sigma\).}
    \label{fig:boxplot_extrapole} 
\end{figure}

In Fig.\ref{fig:boxplot_extrapole} we compare the prediction accuracy across the three datasets, showing that, as a data-driven model, the more we extrapolate from the training space, the worse the inference will be. The best results from the rollout are consistently obtained with trained geometries and loads. In contrast, error accumulates progressively in models facing different load conditions and meshing; however, this growth does not spike for any of the three defined state variables. These findings align with the behaviors observed in the literature and highlight the robustness of graph networks, emphasizing their potential in practical precision medicine applications dependent on digital twins.

As an additional resource, Appendix \ref{an:4l} outlines the network performance across ten simulations conducted at four distinct load points for an untrained anatomy. These load conditions, which differ from those encountered during training, serve as a confirmation of the actuator's adaptability and robustness.

\section{Conclusions}

When neural networks use data-based loss functions for supervised learning, inference is inevitably subject to some degree of bias. Balancing the amount of data needed to ensure model robustness and accuracy in extrapolation remains a complex challenge. However, hybrid methodologies have proven to be more effective than purely data-driven models, particularly in environments with sparse or limited data \cite{TIGNNs, Tierz2024b, Urdeitx2024}. By incorporating soft constraints during loss optimization, imposing mathematical structures on the decoding process, or using base distributions to enhance the expressiveness of the latent space—as demonstrated in variational autoencoders \cite{kingma2022autoencodingvariationalbayes}—we hypothesize that neural networks can achieve more robust inference.

Validating this idea through computational mechanics, we teach the network to understand the physical behavior of soft tissue and act as a temporal integrator. However, we face the challenge that recurrent inference during the rollout tends to accumulate error. Despite this, the network’s understanding of the underlying physics, augmented by data and noise (discussed in more detail in Appendix \ref{an:noise}), aids in achieving realistic dynamic behavior. Additionally, graph neural networks exhibit impressive extrapolation capabilities when applied to unfamiliar geometries or meshes in finite element simulations, suggesting the model’s ability to grasp energy exchanges between the actuator and the receiver.

Therefore, integrating real-time simulation with precision medicine through deep learning is feasible, paving the way for the medical use of patient-specific digital twins. However, new solutions bring new challenges. In this case, the interpretability of message passing and scalability issues with larger meshes remain as the primary limitations in graph architectures. 

These limitations may be addressed by exploring the impact of message passing on model learning while integrating graph representations with model order reduction or multiresolution techniques \cite{ijcai2019p551,hy2022multiresolutionequivariantgraphvariational} to enhance performance and reduce inference times. On the other hand, the interpretability of physics-based biases is also a fundamental requirement for the acceptance of digital twins, specially for critical applications. This holds true for industrial application scenarios where safety and accuracy are paramount (industrial monitoring, aircraft maintenance, etc.), as well as for medical applications in personalized medicine pipelines (virtual surgery planning, telemedicine diagnostics, etc.). However, interpretability requirements should not hinder or constrain a much needed robustness to unknown differential behaviors.

Therefore, we can conclude that this work provides a consistent roadmap for further reducing the heavy data dependency of supervised deep learning approaches. The improvements achieved through the implementation of hybrid-AI have enabled the training of fast, reliable liver digital twins, paving the way for real-time haptic surgical simulations with a high degree of accuracy.

\section*{Acknowledgements}

This work was supported by the Spanish Ministry of Science and Innovation, AEI/10.13039/501100011033, through
Grants number TED2021-130105B-I00 and PID2023-147373OB-I00, and by the Ministry for Digital Transformation
and the Civil Service, through the ENIA 2022 Chairs for the creation of university-industry chairs in AI, through Grant TSI-100930-2023-1.

This material is also based upon work supported in part by the Army Research Laboratory and the Army Research
Office under contract/grant number W911NF2210271.
The authors also acknowledge the support of ESI Group through the chair at the University of Zaragoza.


\newpage

\appendix

\section{Noise data augmentation}
\label{an:noise}

The rollout's inference can be defined as the recursive prediction of the model’s temporal evolution, where accumulated error typically increases with successive predictions, affecting new outputs of the deep neural network. For this reason, robustness mechanisms are essential to prevent error spikes that could destabilize the model.

In this study, the deliberate inclusion of noise is validated as a tool for data augmentation rather than for filtering/denoising purposes. This concept is established by defining \( \bs{z}^t_i \) as the state of a node at a specific time step, represented by the state vectors \( \bs{q}_i \), \( \bs{v}_i \), and \( \bs{\sigma}_i \),
\begin{equation}
\bs{z}^t_i = \begin{pmatrix} \bs{q}_i \\ \bs{v}_i \\ \bs{\sigma}_i \end{pmatrix},
\end{equation}
where the noise methodology defines ${\bs{z}^{\prime}}^t_i$ as the node representation with added noise. Specifically, for the position vector \( \bs{q}_i \), a stochastic term proportional to the mean of the edge distances \( \bs{\overline{u}}_{ij} \) is included, 
\begin{equation}
\bs{q}'_i = \bs{q}_i + \bs{\overline{u}}_{ij} \epsilon_q n_q \quad \epsilon_{q} \sim \mathcal{U}(-1, 1),
\end{equation}
capturing systematic positional variations within the discretization range and scaled with an hyperparameter $n_q$.

For the velocity vector \( \bs{v}_i \) and the stress vector \( \bs{\sigma}_i \), noise is introduced as a relative perturbation, scaled by a factor \( n_v \), 
\begin{equation}
\bs{v}'_i = \bs{v}_i + \bs{v}_i\epsilon_{v}n_v, \quad \epsilon_{\bs{v}} \sim \mathcal{U}(-1, 1),
\end{equation}
\begin{equation}
\bs{\sigma}'_i = \bs{\sigma}_i + \bs{\sigma}_i\epsilon_{\sigma}n_v, \quad \epsilon_{\sigma} \sim \mathcal{U}(-1, 1).
\end{equation}
These stochastic terms are also uniformly distributed, with distribution denoted as \( \mathcal{U} \), within the interval \([-1, 1]\). Terefore, $z'_i$ can be defined as,
\begin{equation}
{\bs{z}^{\prime}}^t_i((\bs{z}_i,\bs{\overline{u}}_{ij} ,n_q, n_v)) = \begin{pmatrix} \bs{q}'_i \\ \bs{v}'_i \\ \bs{\sigma}'_i \end{pmatrix} = \begin{pmatrix} \bs{q}_i + \bs{\overline{u}}_{ij} \epsilon_q n_q \\ \bs{v}_i + \bs{v}_i\epsilon_{v}n_v \\ \bs{\sigma}_i + \bs{\sigma}_i\epsilon_{\sigma}n_v \end{pmatrix},
\end{equation}
\noindent which consequently generates that $\frac{dz_i}{dt}$ denotes the temporal variation of the state variables, but taking into account the variation aggregated by the noise,

\begin{equation}
    \frac{d\bs{z}^{\text{GT}}_i}{dt} =  \frac{\bs{z}^{t+1}_i - \bs{z}^{\prime t}_i}{dt},  
\end{equation}
so that naming the network output as $\frac{dz^{\text{net}}_i}{dt}$, the loss function for the data and consequently the model optimization will be given by Eq. (\ref{eq:lossdata}). The opposite approach, denoising, would have been not to consider this noise variation in the cost optimization, but only at the network input such that,
\begin{equation}
    \frac{d\bs{z}^{\text{GT}}_i}{dt} =  \frac{\bs{z}^{t+1}_i - \bs{z}^t_i}{dt} .  
\end{equation}

This approach establishes a learning process in which the network learns to construct the integrator from positions that are not always exact, building essential robustness for the recurrent inference during rollout.

Additionally, for this project we have not optimized a deep sweep of hyperparameters to regulate the noise in the three state vectors. However, we have worked with some noise values that have proven to improve the inference, but without being able to ensure that they are optimal. These values are $10^{-4}$ for $n_q$ and $10^{-3}$ for $n_v$.

\section{Model scalability and message passing efficiency}
\label{an:12p}

\begin{figure}[p]

 \centering \includegraphics[width=0.8\linewidth]{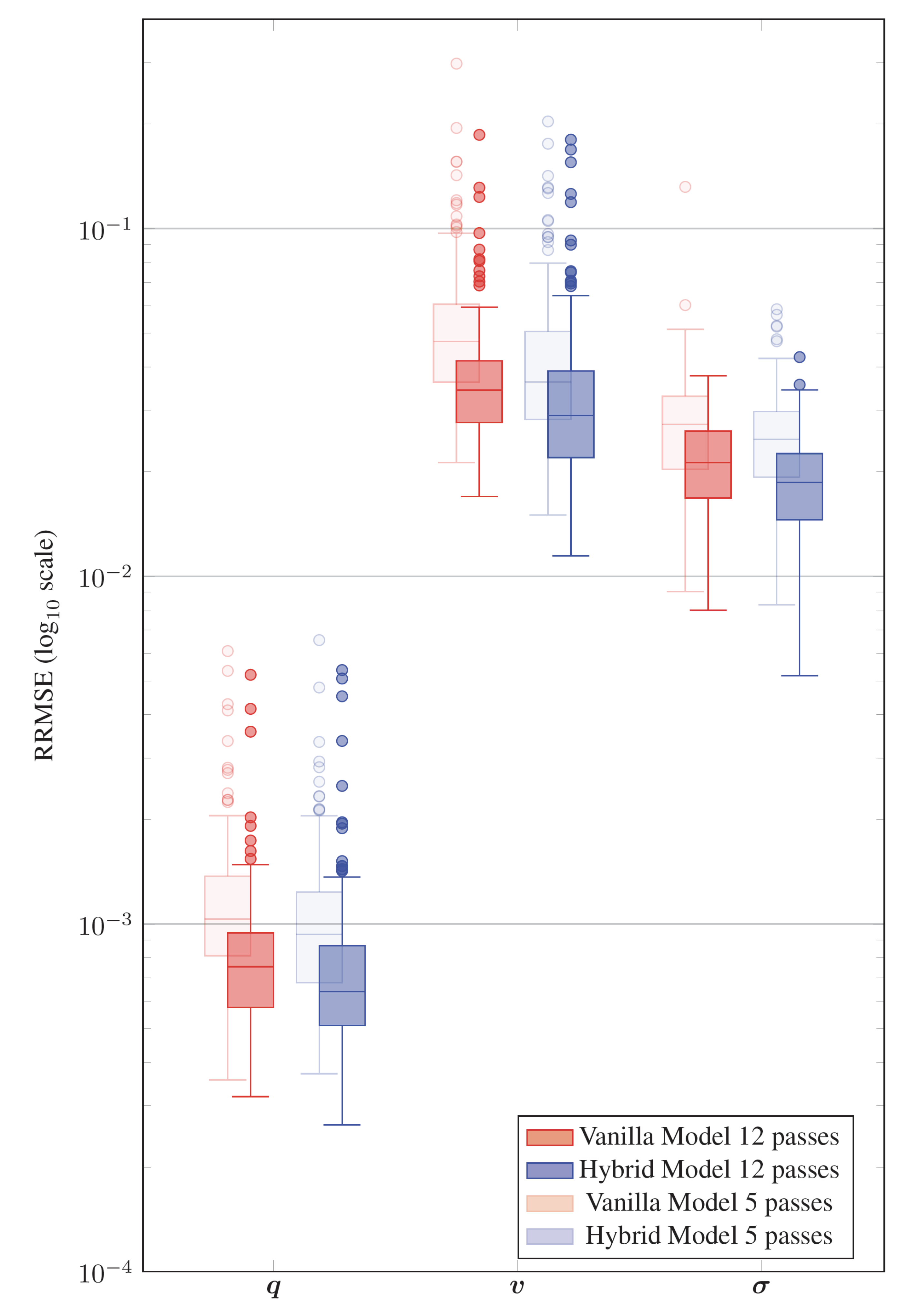}
    \caption{Comparison of RRMSE rollouts between the five-pass and twelve-pass model across 190 unseen simulations in a new anatomy. The blue boxes represent the error distribution for the vanilla model, while the red boxes show the distribution for the hybrid model. The boxes with reduced opacity illustrate the error distributions previously displayed in the Fig. \ref{fig:boxplot_20k_combined}, serving as a visual aid for comparison.} 
    \label{fig:boxplot_2m_combined} 
\end{figure}

\begin{figure}[h!]
\centering \includegraphics[width=\linewidth]{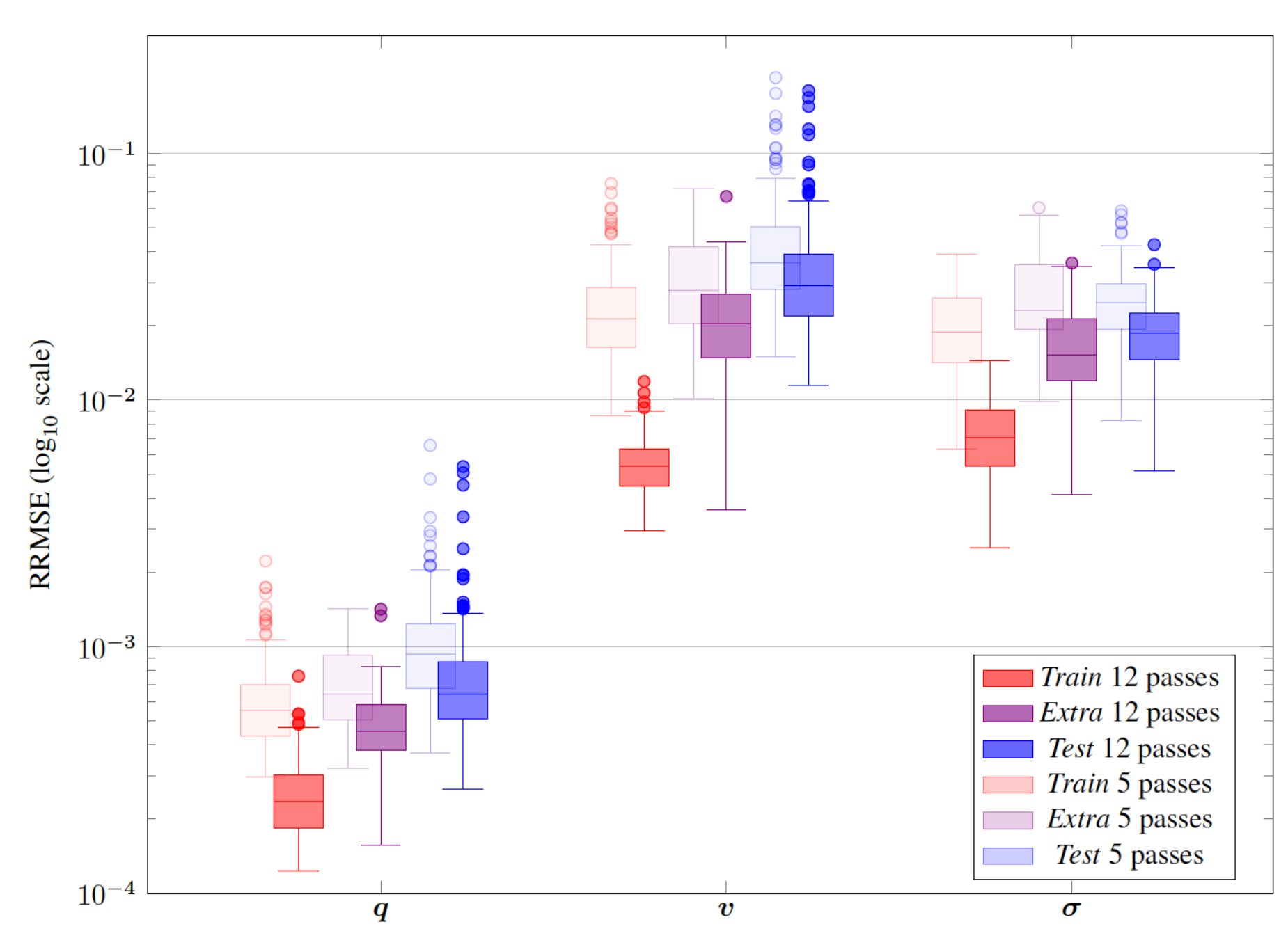}
    \caption{Comparison of RRMSE rollouts between the five-pass and twelve-pass hybrid models across different datasets: \textit{Train} (trained data), \textit{Extra} (unseen load states with known geometry), and \textit{Test} (untrained geometries and load states, main testing dataset) for the parameters \(\bs q\), \(\bs v\), and \(\bs \sigma\). The boxes with reduced opacity represent the error distributions from the five-pass results previously shown in the Figure \ref{fig:boxplot_extrapole}, serving as a visual aid for comparison.} 
    \label{fig:boxplot_2m_combined2} 
\end{figure}

As we have observed, increasing the steps in the graph processor scales almost linearly for both training and inference. However, even with the proper distribution of information provided by the actuator graph, extending each node’s attention to its surroundings significantly reduces error, as demonstrated in Fig. \ref{fig:boxplot_2m_combined}, especially when considering the previously presented error in Fig.\ref{fig:boxplot_20k_combined}. This effect is not only amplified in the inference on the main \textit{Test} database , but it is also significantly correlated in the rollout error on the other datasets ({\textit{Train}} and {\textit{Extra}}, Fig. \ref{fig:boxplot_2m_combined2}).

This main hypothesis is based on how each node infers a response that is mathematically dependent on the behavior of further nodes, which provides greater stability to the system without being significantly affected by an increase in the dimensionality of the latent space. While a higher-dimensional latent space aids in better projecting information related to the state variables, the weight of the inference seems to depend more on the proper optimization of message passing.

\begin{figure}[h!]
    \centering
    \includegraphics[scale=0.8]{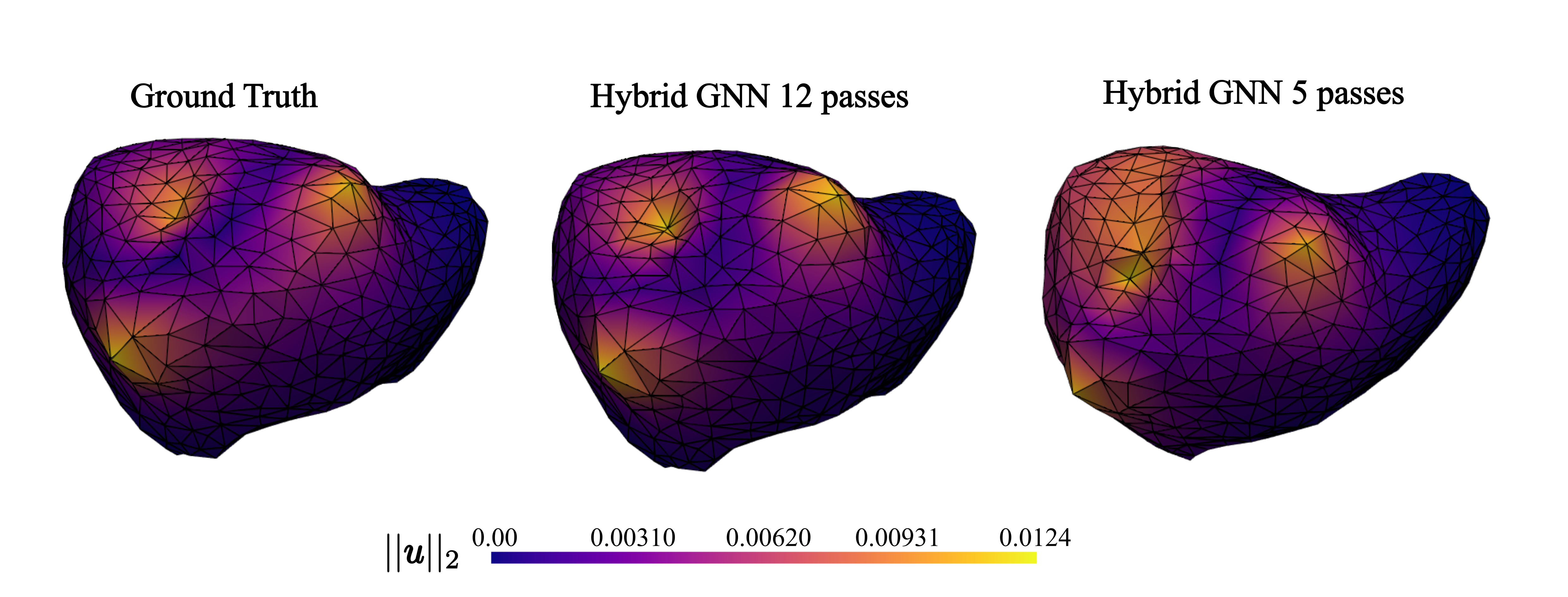}
    \caption{Comparison of the final inference step between the two proposed models and the numerical solution, used as ground truth, for untrained geometries and unseen load conditions. The color scale represents the 2-norm of the displacement vector on each node.}
    \label{fig:comp3}
\end{figure}

Accordingly, in Fig. \ref{fig:comp3} we observe how, for the same test simulation, the model with a greater number of passes better captures the locality of the deformation, preventing its propagation and demonstrating a deeper understanding of the soft tissue properties.

\begin{figure}[h!]
\centering \includegraphics[width=\linewidth]{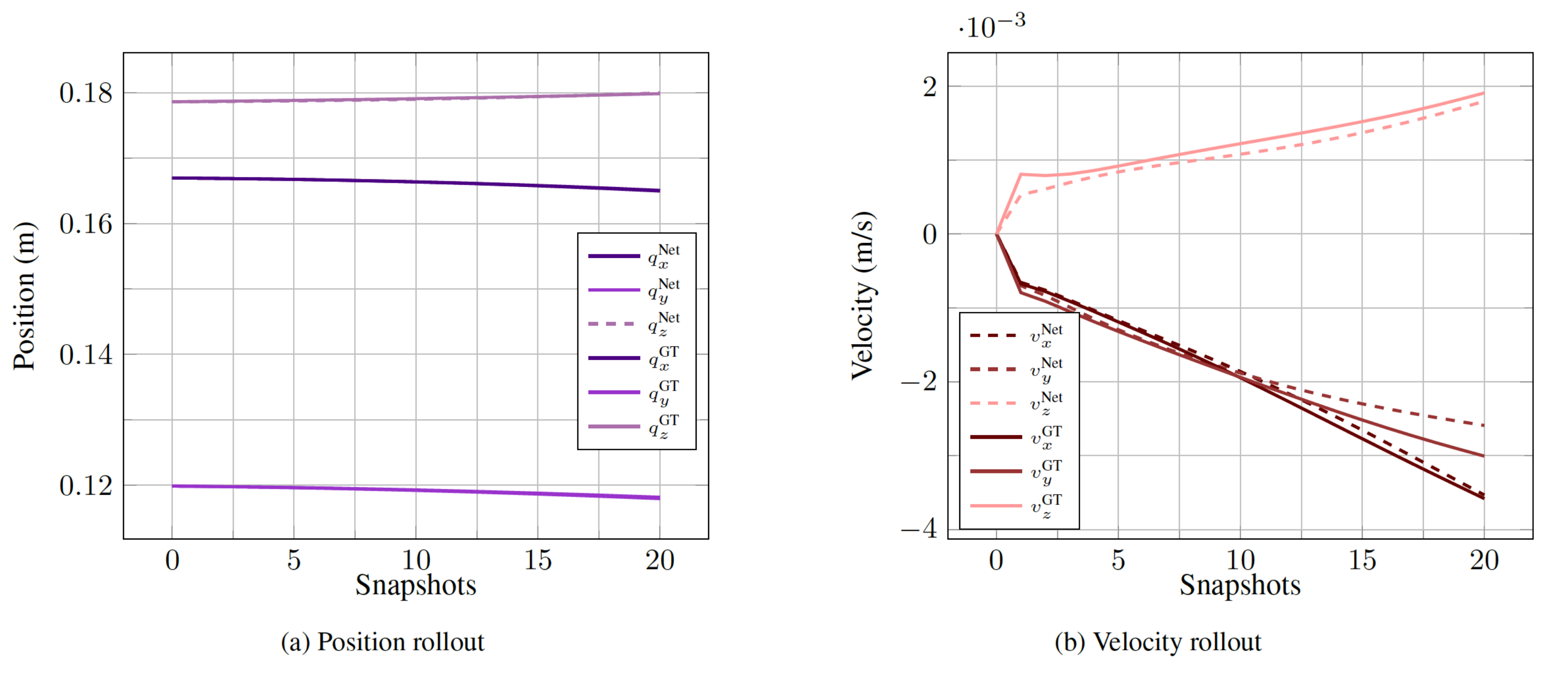}
    \caption{Comparison of the rollout for first six state variables at a randomly selected node within one of the test inferences for the twelve-pass hybrid model. The dotted lines represent the network's recurrent prediction compared to the ground truth from numerical simulation, illustrating the accumulation of relative error over the course of temporal evolution.}
    \label{fig:subplot_mosaic1}
\end{figure}

\begin{figure}[h!]
    \centering

\centering \includegraphics[width=\linewidth]{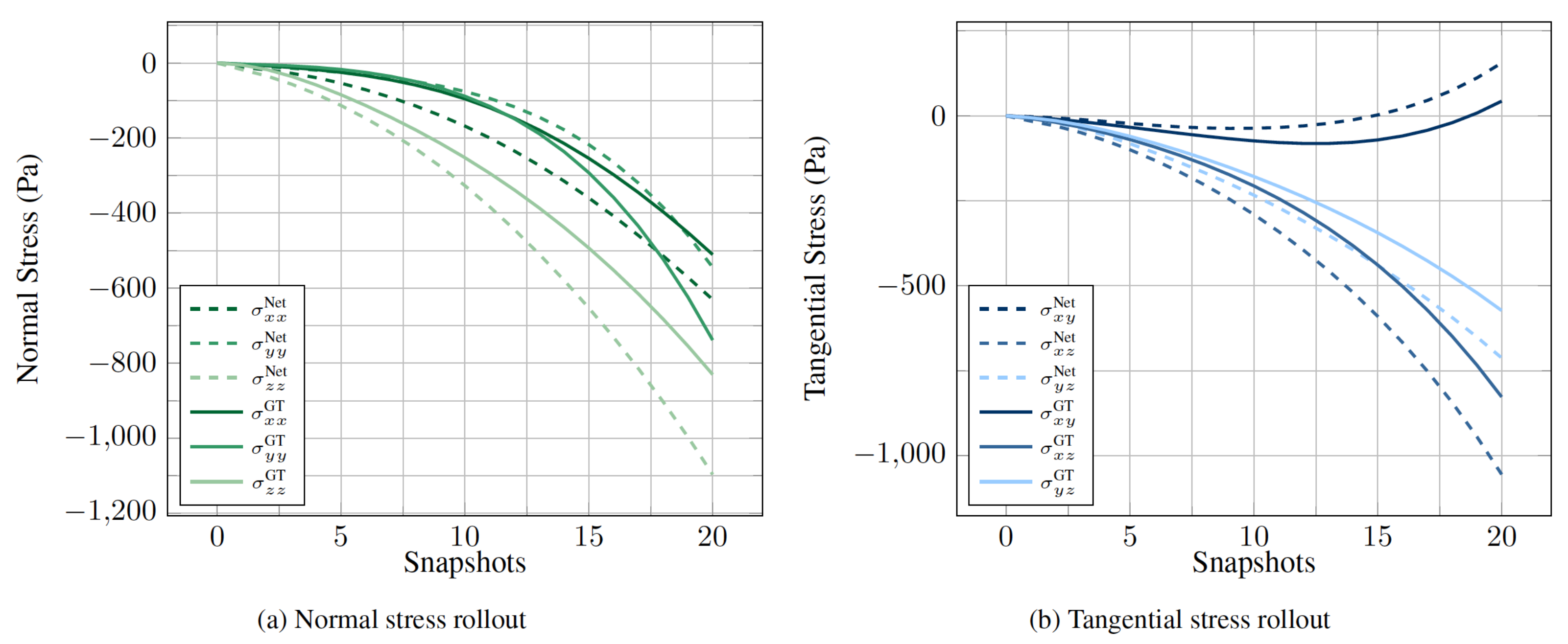}
    \caption{Comparison of the rollout for last six state variables at a randomly selected node within one of the test inferences for the twelve-pass hybrid model. The dotted lines represent the network's recurrent prediction compared to the ground truth from numerical simulation, illustrating the accumulation of relative error over the course of temporal evolution.}
    \label{fig:subplot_mosaic2}
\end{figure}

As supplementary material, the temporal evolution of the twelve state variables is also presented in four different subplots, shown in Figs. \ref{fig:subplot_mosaic1} and \ref{fig:subplot_mosaic2}. These illustrate the network's adjustment for inferring a node in an untrained geometry and load state. This analysis clearly validates how recurrent inference accumulates errors due to temporal integration. However, thanks to the appropriately biased methodology, along with the robustness achieved through data augmentation via noise, the simulation is prevented from diverging into significantly larger errors.

\section{Time response evolution with passing augmentation}
\label{an:time}

\begin{figure}[h]
\centering \includegraphics[width=\linewidth]{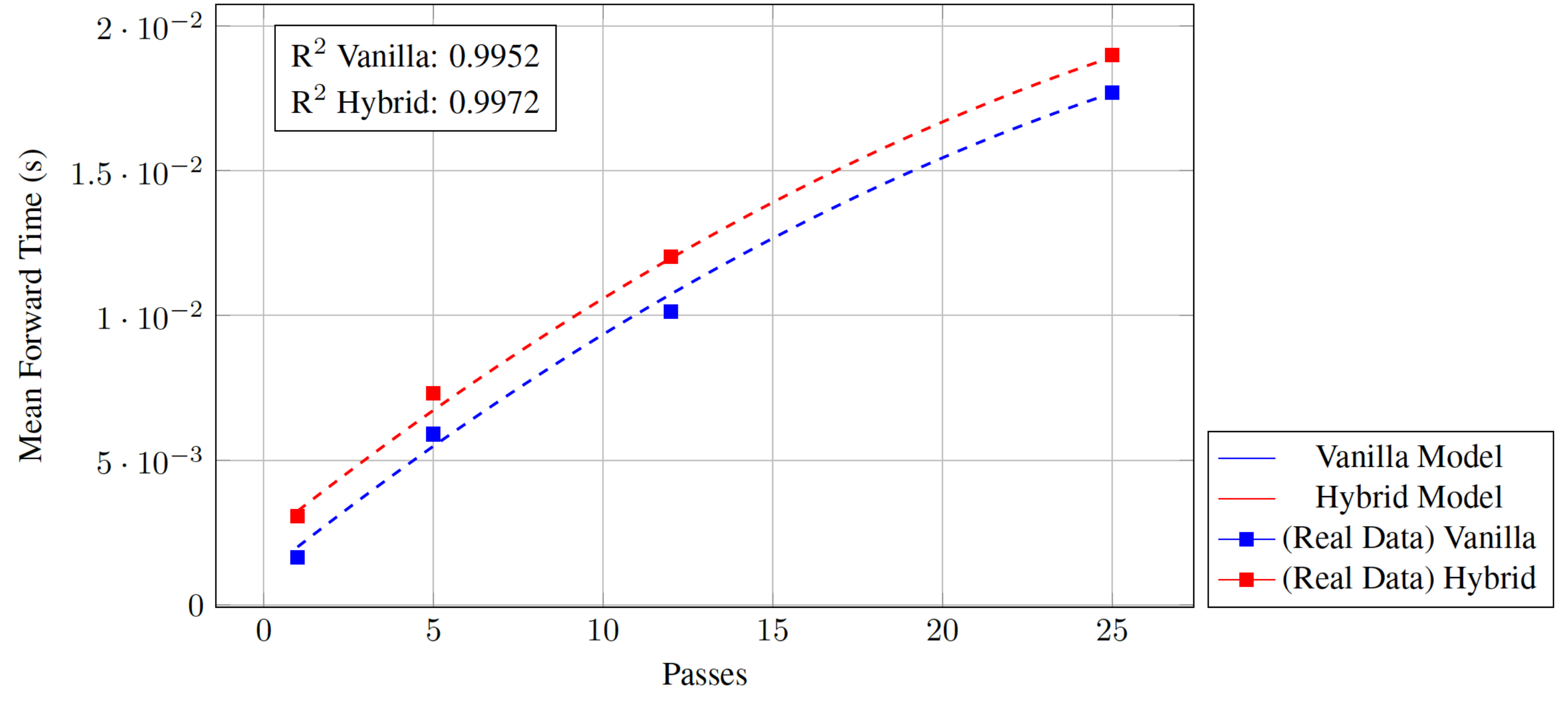}
    \caption{Evolution of response time as a function of the number of steps for both the hybrid and vanilla models. The curve represents a quadratic approximation of the function defined by the test points on a RTX3090 with the $\text{R}^2$ error of each fit.}
    \label{fig:timeEvol}
\end{figure}

In Graph Neural Networks (GNNs), model performance is strongly influenced by the flow of messages, which is controlled by the number of hyperparameter-tuned passes through the graph processing layers. When information is distributed and the geometric function depends on a broader attention area, inference tends to be more stable. However, this improvement comes at the cost of increased computational cost for both backpropagation and inference, as it is a non-parallelizable process. Fig. \ref{fig:timeEvol} illustrates the evolution of response time as a function of the number of steps on a Nvidia RTX3090, for both the hybrid model and the vanilla model.

The time response functions for both the hybrid model and the vanilla model can be regressed to quadratic equations of the form:
\begin{equation}
    T(x) = ax^2 + bx + c, \quad x \in \mathbb{N},\; \{a,b,c\}  \in \mathbb{R}
\end{equation}
where:
\begin{itemize}
    \item \( T(x) \) represents the response time,
    \item \( x \) denotes the number of hyperparameterized steps in the graph processor, where \( x \in \mathbb{N} \).
    \item \( a \) and \( b \) are coefficients that influence the shape and slope of the quadratic curve, and
    \item \( c \) is a constant term that represents the baseline response time when no steps are taken.
\end{itemize}

Given the average coefficients for \( a \) and \( b \):
\[
a \approx -1.082 \times 10^{-5}, \quad b \approx 9.346 \times 10^{-4},
\]
we observe that both models exhibit similar slopes. This parallelism confirms us that these coefficients are shared between the models, while only varying the coefficient \( c \), which directly impacts the baseline response time.

For the two models, the coefficients \( c \) are as follows:
\begin{itemize}
    \item {Coefficient \( c \) for vanilla}: \( c_{\text{vanilla}} = 0.00108, \)
    \item {Coefficient \( c \) for hybrid}: \( c_{\text{hybrid}} = 0.00232. \)
\end{itemize}

In summary, the coefficients \( a \) and \( b \) determine the curvature and initial slope of the response time functions, primarily related to the computational weight of the loop associated with the graph processor. Meanwhile, the coefficient \( c \) serves as a differentiator, representing the weight of the decoder in the response. The difference of \( 0.00124 \) seconds, associated with the nodal reparameterization and the aggregation of boundary contributions, highlights the trade-off between stability and computational efficiency.

\section{Multi-graph contribution to message passing}
\label{an:mssg}

In the geometric context of graph networks, the discussion focuses on how message transmission ---determined by the number of processing steps--- positively influences the local understanding of the system’s mechanical behavior. However, we must take into account that broader message distribution incurs a high computational cost for both inference and backpropagation, reopening the debate between performance and efficiency.

Single-graph models \cite{TIGNNs, Tierz2024b} suffer significantly from this limitation: if a local load is communicated relatively far away (at higher geodesic distances \cite{Bronstein-2021_geodistnce}), the message flow will not inform more distant nodes, a limitation that becomes more pronounced with finer spatial discretization. In this context, the proposed multigraph scheme \cite{pfaff2021learningmeshbasedsimulationgraph}—validated in Appendix \ref{an:mssg}—has a clear objective: to improve information distribution. From the initial steps, it conveys not only where energy is introduced into the system but also the relative distance from each node.

For this project, the single-graph model appends the actuator information $u_i$ node-wise during message transmission, which limits its influence on nodes farther from the radius defined by the message-passing modules hyperparameter $\mathcal{M}$,
\begin{equation}\label{eq:Processor2}
    f^{m}(\bs{x}^{m}_{ij}, \bs{x}^m_i, \bs{x}^m_j) \rightarrow \bs{x}'^{m}_{ij}, \quad f^{v} \left(\bs{x}_i, \phi (\bs{x}'^{m}_{ij}), \bar{\bs u}_i ) \right) \rightarrow \bs{x}'_{i},
\end{equation}
where $(\cdot, \cdot, \cdot)$ denotes vector concatenation and $x'_i$ and $x'_{ij}$ are the updated nodal and edge latent representations for the mesh-graph.

\begin{figure}[h!]
\centering \includegraphics[width=\linewidth]{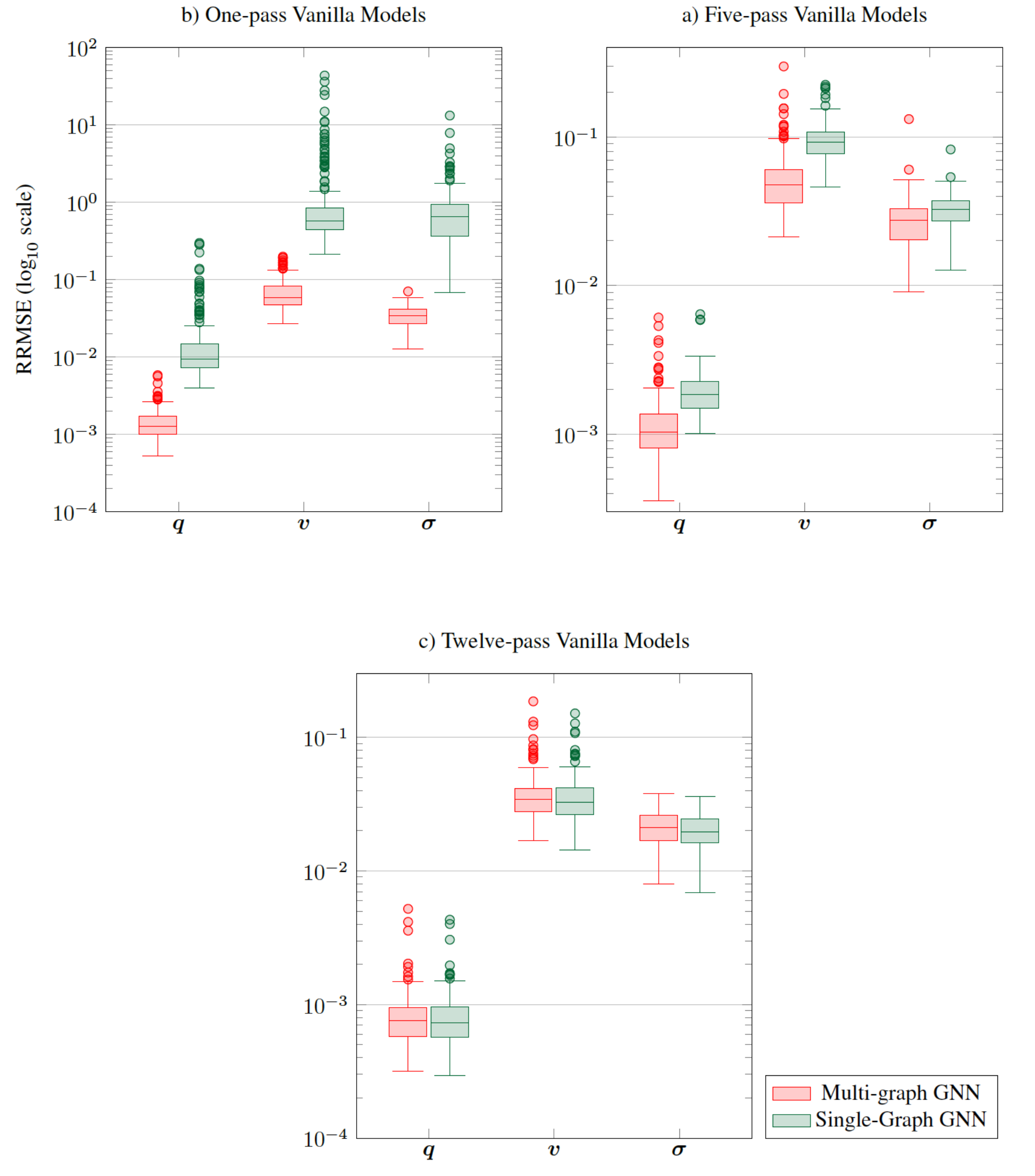}
    \caption{Relative error comparison of data-driven model performance for single-graph versus multi-graph frameworks within the processor, under equivalent training conditions. The left panel displays results for both models with a single pass transmission, the right panel shows results for five passes, and the bottom panel presents results for twelve passes.} 
    \label{fig:mssg1} 
\end{figure}

Fig. \ref{fig:mssg1} compares the relative error between single-graph models and the multigraph model, supporting the hypothesis that enhanced information transmission improves inference accuracy in scenarios where message passing is limited in relation to the density of the mesh. This comparison is performed under identical parameter, training, and testing conditions, validating the choice to adopt a multigraph approach. The results indicate that this approach not only enhances model performance but may also improve the efficiency of message flow within the graph processor.

For models trained with only five passes Fig. \ref{fig:mssg1} (a) and Fig. \ref{fig:mssg1} (b), the multigraph scheme significantly improves information flow. However, when the number of passes is sufficient to cover the entire spatial discretization, see Fig. \ref{fig:mssg1} (c), the multigraph effect is less critical to the network’s performance. This may indicate that after twelve steps, the 500-node graph reaches a stationary state of information distribution, in which all nodes are theoretically aware of the external excitation affecting the system, and the inference performance is not constrained by it.

\section{Performance under unknown loading states}
\label{an:4l}

For the training database, we chose that the number of loaded nodes range from a minimum of one to a maximum of three. In practice, endoscopic surgery uses a maximum of two surgical tools (a third incision could be made for the camera). 


This appendix provides an overview of the models' performance under extrapolation conditions, more challenging than those encountered during the initial tests. Meanwhile we aim to validate the robustness and adaptability of the multi-graph networks approach proposed.

The test set will consist of 10 simulations with four imposed displacement (traction or compression) application points. This will result in a system comprising four actuator graphs and one central graph, subsequently denoted as Extra*.

Table \ref{tab:combined} presents a comparison of the absolute and relative errors between the two methodologies, highlighting their comparable robustness. This similarity is likely attributed to the shared geometric bias of both models, which primarily equips the network with effective generalization capabilities.

\begin{table}[h!]
    \centering 
    
    \begin{minipage}[t]{0.48\textwidth}
        \centering
        \begin{tabular}{@{}lcc@{}}
            \toprule
            & \textbf{Hybrid} & \textbf{Vanilla} \\ \midrule
            $\bs{q}$ (m): & $3.620 \times 10^{-4}$ & $3.58 \times 10^{-4}$ \\
            $\bs{v}$ (m/s): & $6.16 \times 10^{-4}$ & $6.23 \times 10^{-4}$ \\
            $\bs{\sigma}$ (Pa): & $7.236\times 10^{2}$ & $7.562 \times 10^{2}$ \\ 
            \bottomrule
        \end{tabular}
    \end{minipage}
    \hfill
    \begin{minipage}[t]{0.48\textwidth}
        \centering
        \begin{tabular}{@{}lcc@{}}
            \toprule
            & \textbf{Hybrid} & \textbf{Vanilla} \\ \midrule
            $\bs{q}$: & $1.2717 \times 10^{-3}$ & $1.203 \times 10^{-3}$ \\
            $\bs{v}$: & $5.13 \times 10^{-2}$ & $5.18 \times 10^{-2}$ \\
            $\bs{\sigma}$: & $3.28 \times 10^{-2}$ & $3.40 \times 10^{-2}$ \\ 
            \bottomrule
        \end{tabular}
    \end{minipage}
    \vspace{0.4cm}
    \caption{Side-by-side comparison of RMSE and RRMSE across hybrid and vanilla models for all snapshots in the testing subset. Each table evaluates five passes using approximately 2 million trainable parameters, with twelve passes applied to obtain the best-performing weights. The testing subset consists of 10 previously unseen simulations in a new anatomy featuring four loading points, $\bs{L5}$.}
    \label{tab:combined}
\end{table}

\begin{figure}[h!]
    \centering
    \includegraphics[scale=0.8]{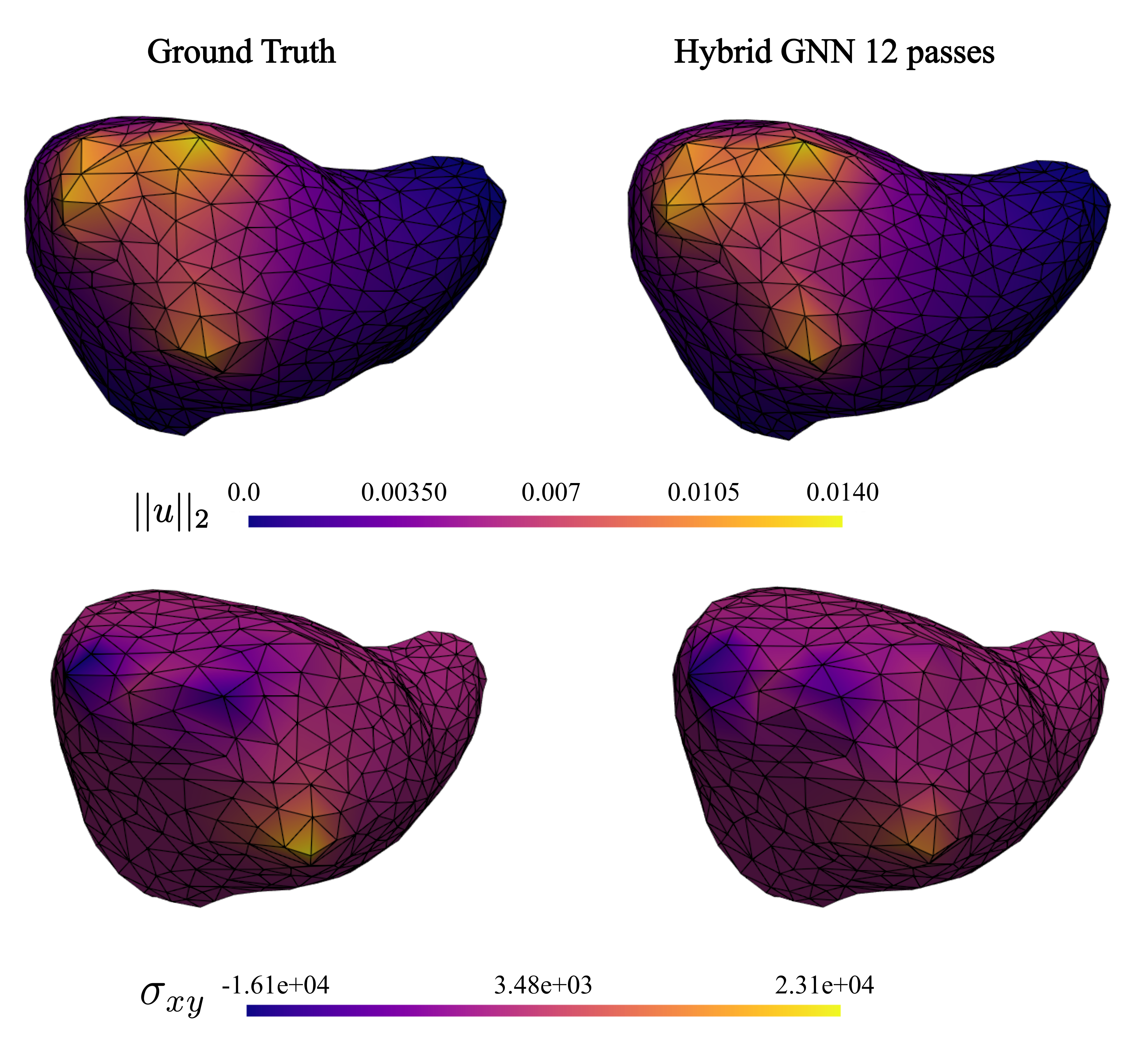}
    \caption{Comparison of the final inference step between the model's prediction and the numerical solution, used as the ground truth, for untrained geometries with four imposed displacements. The color scale represents the 2-norm of the displacement vector at each node for the top comparison, while for the bottom comparison, it shows the 2-norm  of the tangential stresses. Both comparisons correspond to two separate test simulations, each with four imposed displacements, in a twelve-pass model trained with a maximum of three imposed actuators.}
    \label{fig:comp31}
\end{figure}

On the other hand, Figure \ref{fig:comp31} illustrates the predictions for the final step of the rollout in two distinct simulations from the test set. These simulations correspond to a displacement field and a tangential stress field, respectively, highlighting the network's capabilities.

\begin{figure}[h]
\centering \includegraphics[width=\linewidth]{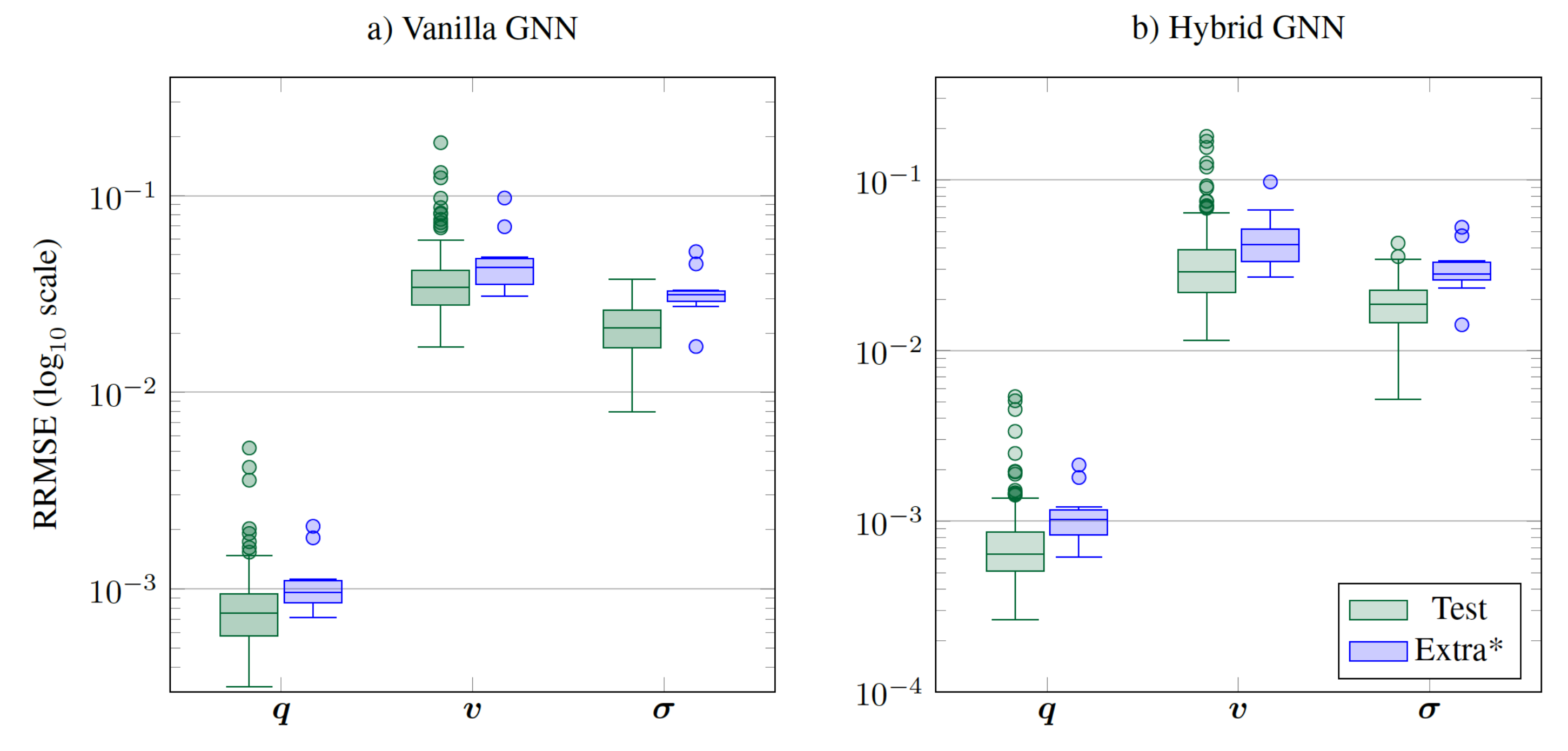}
    \caption{Comparison of RRMSE rollouts between the five-pass and one-pass models across 190 unseen simulations in a new anatomy. The blue boxes represent the error distribution for the vanilla model, while the red boxes show the distribution for the hybrid model.} 
    \label{fig:4loads-box} 
\end{figure}

Finally, the boxplot presented in Fig. \ref{fig:4loads-box} compares the progression of the relative error during extrapolation to a more complex dataset against the primary test set (referred to as \textit{Test}).

\end{document}